\documentclass[10pt,twocolumn]{article}

\usepackage[letterpaper,margin=0.75in]{geometry}

\usepackage[hyphens]{url}  
\usepackage{graphicx} 
\usepackage{caption} 

\usepackage[numbers,sort&compress]{natbib}
\usepackage{hyperref}

\usepackage{booktabs}
\usepackage{multirow}
\usepackage{array}
\usepackage{tabularx}
\usepackage{makecell}

\newcolumntype{Y}{>{\centering\arraybackslash}X}

\newcommand{\ph}[1]{{\normalfont\itshape$\langle$#1$\rangle$}}
\newcommand{\promptcard}[2]{
  \par\smallskip\noindent
  \tikz\node[draw=black!15, line width=0.4pt, rounded corners=5pt, fill=black!4,
    inner sep=6pt, text width=\dimexpr\linewidth-14pt\relax, align=left]{%
    \footnotesize{\scshape #1}\\[3pt]{\ttfamily\footnotesize #2}};%
  \par\smallskip}
\usepackage{amsmath}
\usepackage{amssymb}

\usepackage{tikz}
\definecolor{featink}{HTML}{6E6E6E}
\newcommand{\featpic}[1]{\mbox{\begin{tikzpicture}[baseline=-0.6ex,line join=round]#1\end{tikzpicture}}}
\DeclareRobustCommand{\gH}{\featpic{\path[draw=featink,fill=white,line width=0.5pt] (-0.24em,-0.18em)--(0.24em,-0.18em)--(0,0.26em)--cycle;}}
\DeclareRobustCommand{\gP}{\featpic{\path[draw=featink,fill=white,line width=0.5pt] (0,-0.24em)--(0.22em,0)--(0,0.24em)--(-0.22em,0)--cycle;}}
\DeclareRobustCommand{\gR}{\featpic{\draw[draw=featink,fill=white,line width=0.5pt] (0,0.02em) circle[radius=0.22em];}}
\DeclareRobustCommand{\gA}{\featpic{\draw[draw=featink,line width=0.45pt] (-0.22em,-0.20em) rectangle (0.22em,0.24em); \draw[draw=featink,line width=0.45pt] (0,-0.20em)--(0,0.24em); \draw[draw=featink,line width=0.45pt] (-0.22em,0.02em)--(0.22em,0.02em);}}
\DeclareRobustCommand{\gS}{\ensuremath{\ast}}
\DeclareRobustCommand{\lb}{{\color{featink}[}}\DeclareRobustCommand{\rb}{{\color{featink}]}}
\DeclareRobustCommand{\lp}{{\color{featink}(}}\DeclareRobustCommand{\rp}{{\color{featink})}}

\usepackage{algorithm}
\usepackage{algorithmic}

\usepackage{newfloat}
\usepackage{listings}
\DeclareCaptionStyle{ruled}{labelfont=normalfont,labelsep=colon,strut=off} 
\floatstyle{ruled}
\newfloat{listing}{tb}{lst}{}
\floatname{listing}{Listing}

\usepackage{booktabs}

\title{Locating Answer-Correctness Signals in Frozen Large Language Models}

\author{Yuansen Liu,\ \ Yixuan Tang \thanks{Corresponding author.},\ \ Anthony Kum Hoe Tung \\
School of Computing, National University of Singapore \\
\{yuansen, yixuan, atung\}@comp.nus.edu.sg
}
\date{}

\begin{document}

\maketitle
\begin{abstract}
Language models expose internal signals that predict whether an answer is correct, readable from a single forward pass of a frozen model without additional generations. Yet existing probes often commit to one signal family or layer and can be brittle under distribution shift; in retrieval-augmented settings, many specialized detectors instead target passage faithfulness, which can diverge from correctness when retrieved evidence is unhelpful or conflicting. We therefore ask where answer correctness is readable, which internal signal families carry it, and how they should be combined. We search over hidden states, token probabilities, residual-stream features, attention, and their fusion, treating the selected readouts as a predictive measurement rather than a mechanistic localization. We run this analysis separately in closed-book and with-context settings, since context can change which readouts are informative. A consistent anatomy emerges: correctness concentrates in the answer span, recovered from the answer tokens even under retrieval, and the families carry it complementarily, so fusing them helps most out of distribution, where a single signal is weakest. The protocol is effective across two backbones and gates a retrieval controller as one downstream use.
\end{abstract}
 
\section{Introduction}

Large language models encode, in their own internal states, a signal of whether the
answer they are about to give is correct. Hidden states, token-level probabilities,
residual-stream norms, and attention distributions all separate correct from incorrect
generations, often more reliably than anything
the model says out loud \citep{azaria2023internal,kossen2024semantic,zhang2025icr}, and
reading it from a single forward pass turns a frozen model into a
correctness detector.

\begin{figure}[t]
\centering
\includegraphics[width=\columnwidth]{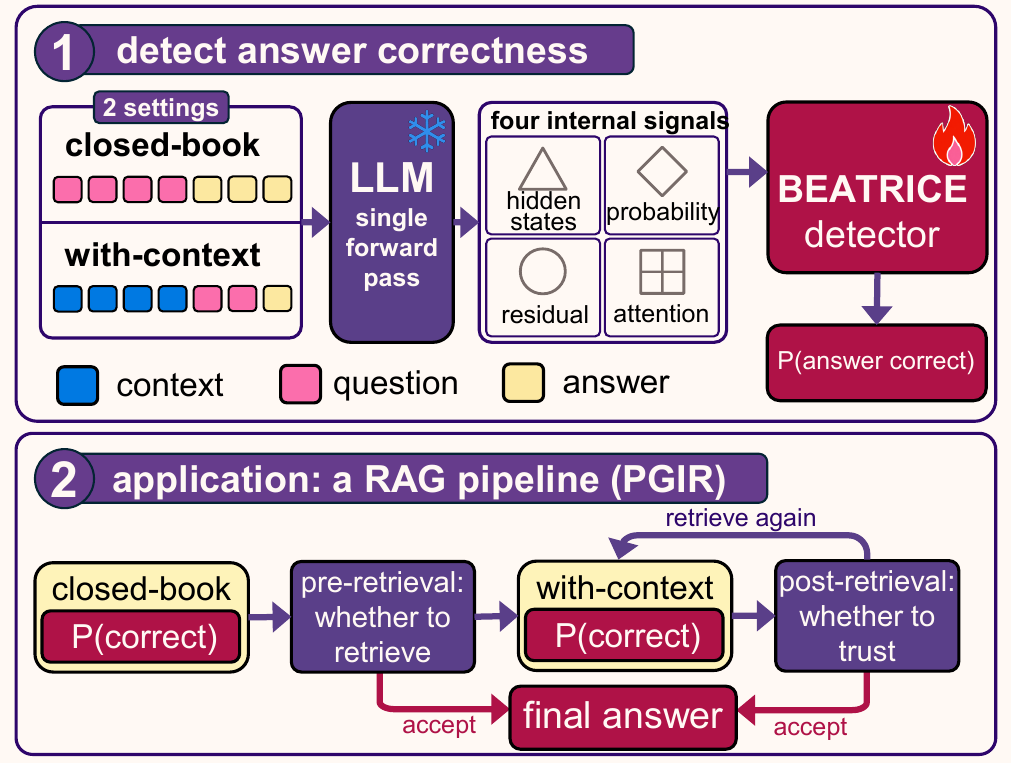}
\caption{This work in one view. Top: in either setting, one forward pass of the frozen
LLM yields four families of internal signals, which the BEATRICE detector reads into a
probability that the answer is correct. Bottom: in the PGIR pipeline the closed-book
estimate decides whether to retrieve at all, the with-context estimate whether to
trust the grounded answer or retrieve again.}
\label{fig:intro}
\end{figure}

Yet the internal correctness signal is less settled than its headline results suggest. Most
probes read a single signal at a single layer, chosen by convention rather than analysis
\citep{azaria2023internal,binkowski2025hallucination}, and recent re-evaluations find such
detectors brittle and their reported gains fragile under careful testing
\citep{janiak2025illusion}. Three questions remain open: where the correctness signal
can be read out, what kind of signal carries it, whether hidden states, output
probabilities, residual-stream norms, or attention distributions, and how it should be read
out and combined rather than assumed linearly decodable from a single site. How these
questions are answered bears on whether a detector holds up when the deployment
setting changes.

Whether the model answers from its own parametric knowledge or from a retrieved passage is
the condition that matters most for this signal, so we study two settings we call
\textit{closed-book} and \textit{with-context}. Conditioning an answer on a passage changes
what the model attends to, what its token probabilities mean, and which of its internal
signals are informative, and it can raise a model's chance of answering correctly or lower it
\citep{mallen2023not,cuconasu2024power}, so a cue learned closed-book has no reason to read
correctness the same way once a passage is in the prompt \citep{baek2025probing}. Placing
the two settings side by side lets us ask whether the anatomy of the correctness signal is
stable across them or specific to each, rather than assume that what is learned in one
transfers to the other.

We instantiate this analysis with \textbf{BEATRICE} (\textbf{B}undled \textbf{E}vidence from
\textbf{A}ttention, \textbf{T}oken-probabilities, \textbf{R}esidual-streams, and
\textbf{I}nternal states for \textbf{C}orrectness \textbf{E}xamination), which reads all four
families, hidden states, token-probability statistics, residual-stream norms, and attention
distributions, from a single forward pass of a frozen language model, and is built by
searching over which signals to read and how to fuse them (Figure~\ref{fig:intro}). Because
the closed-book and with-context forwards are searched separately, the search does double
duty: beyond the detector, it reads out which signals carry correctness in each
setting and how the settings relate.

The same read-out also has a use. Because BEATRICE scores the model both before and after a
passage is retrieved, its closed-book estimate can decide whether retrieval is worth running
and its with-context estimate whether to trust the grounded answer. We connect the two into
a controller, Probe-Gated Iterative Retrieval (\textbf{PGIR})
\citep{jeong2024adaptive,su2024dragin,baek2025probing}, and report it as one downstream
demonstration rather than the point of the paper.

\paragraph{Contributions.}
\begin{itemize}
\item \textbf{An anatomy of the correctness signal.} Reading four families of internal
signal together, we map how answer correctness can be read out: one family dominates while the
others carry it to differing degrees, the usable signal concentrates in the answer span rather than
the question or context, and what gains there are come from combining signals that are
complementary rather than redundant. We characterize this in both the closed-book and
with-context settings and show how the two relate.
\item \textbf{A read-out that holds up.} The same search yields BEATRICE, a detector that
reads correctness from a single forward pass and stays accurate in-domain and zero-shot on
three out-of-distribution datasets, where recent re-evaluations warn that such probes often
fail to hold up \citep{janiak2025illusion}.
\item \textbf{A downstream application.} We turn the read-out into PGIR, a controller that
reads the model closed-book to decide whether to retrieve, with context whether
to trust the result \citep{jeong2024adaptive,su2024dragin}.
\end{itemize}

\section{Related Work}

\paragraph{White-box detection from internal states.}
Reading a frozen model's internal activations to judge whether its
output is correct has become a standard white-box approach, and most
such probes attach to a single signal family. Hidden states are the
most common target: \citet{azaria2023internal} classify layer
activations, INSIDE reads their covariance \citep{chen2024inside}, and
later work tracks their layerwise dynamics or learns from unlabeled
generations \citep{zhang2025icr, ICLR2025_a712d461, kossen2024semantic,
su2024mind, du2024haloscope}. A parallel line reads attention maps
\citep{chuang2024lookback} or their spectral statistics
\citep{binkowski2025hallucination}.
Output distributions supply a third signal, through logit-level checks
\citep{sriramanan2024llm} or a model's own probability that its answer
is correct \citep{kadavath2022}, though a recent re-evaluation finds
single-signal probes brittle across settings \citep{janiak2025illusion}.
Closest to us, a few methods ensemble more than one internal signal:
MultiHaluDet probes hidden states together with multi-scale attention
\citep{alvi2026multihaludet}, and EnsemHalDet stacks hidden-state and
attention detectors for vision-language models
\citep{miyazato2026ensemhaldet}. Both combine at most two families by
late-stage voting and report only that mixing sometimes helps. We unify
four signal families under both early and late fusion, search over their
combinations, and characterize which combinations are complementary
rather than redundant.

\paragraph{Black-box detection from sampling.}
A second family estimates correctness without opening the model, by
drawing several samples and measuring their agreement. SelfCheckGPT
scores cross-sample consistency \citep{manakul2023selfcheckgpt},
semantic-entropy methods cluster samples by meaning
\citep{kuhn2023semantic}, and sequential schemes stop sampling once
confident \citep{wang2023hallucination}. These methods need many
decodes per query, whereas ours reads a single forward pass and
adds no generation cost.

\paragraph{Faithfulness and adaptive retrieval.}
When a retrieved passage is present, a related but distinct question is
faithfulness, whether the answer is grounded in the passage rather than
whether it is correct; the two diverge when retrieval is unhelpful or
conflicts with parametric knowledge, or incomplete \citep{mallen2023not,
cuconasu2024power, xu2024knowledge, DBLP:conf/emnlp/TRACER, DBLP:journals/corr/abs-2604-19005}, and recent detectors target
faithfulness through mechanistic or context-knowledge signals
\citep{sun2025redeep, yeh2026lumina, xiong2025toward, hu2026detecting}.
We instead detect correctness in both settings and use the detector to
control retrieval. This connects to adaptive retrieval, which decides when to retrieve from
question difficulty or model uncertainty \citep{jeong2024adaptive,
su2024dragin, yao2025seakr, wang2023self, baek2025probing}, and
concurrent work extends hidden-state control heads to agentic routing
and tool-use decisions \citep{ghasemabadi2026multihead}; our
controller gates both the decision to retrieve and the decision to
trust a retrieved passage on the same correctness probe.


\begin{figure*}[!t]
\centering
\includegraphics[width=\textwidth]{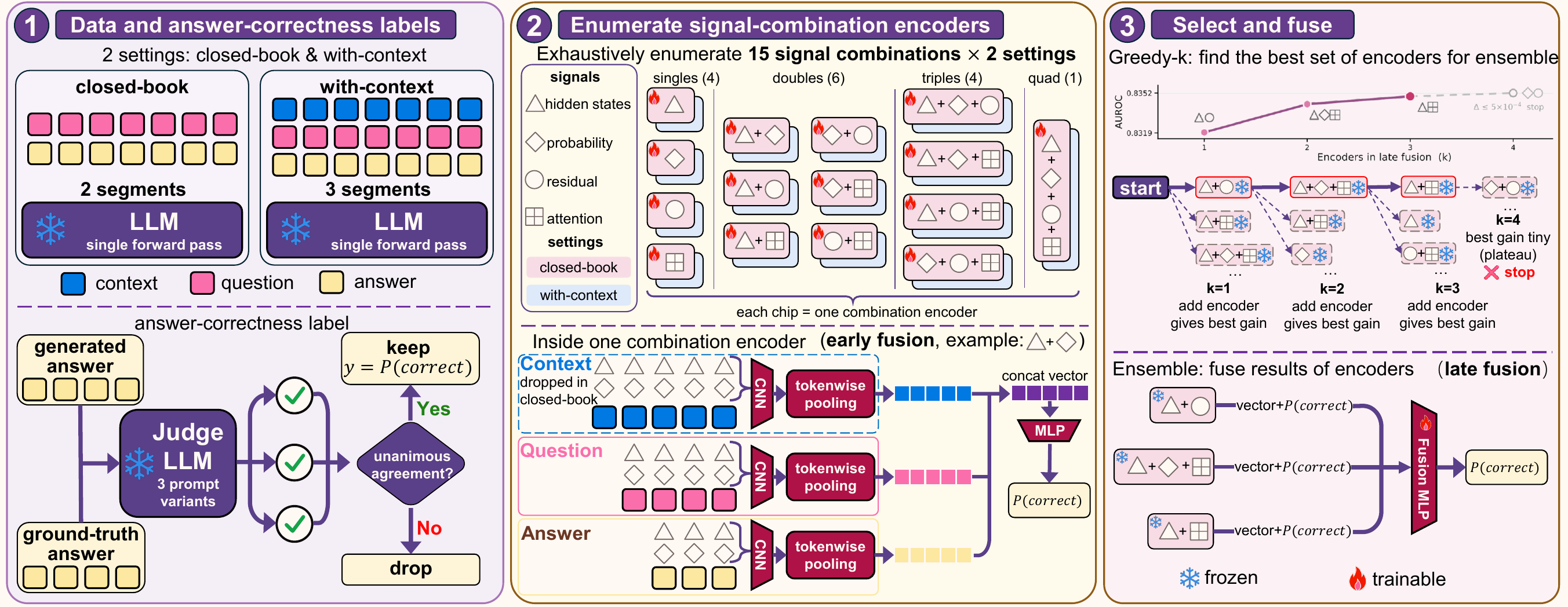}
\caption{The BEATRICE pipeline. (1) Data and labels: the frozen LLM answers once per
setting and a judge LLM keeps only unanimously labeled answers. (2) Combination
encoders: each of the 15 signal-family combinations trains an encoder that pools
token-level signals per segment into a correctness probability (early fusion).
(3) Selection: greedy forward selection grows the ensemble on validation until the gain
plateaus (late fusion).}
\label{fig:overview}
\end{figure*}

\section{Method}

Figure~\ref{fig:overview} gives an overview of the full pipeline, from label
construction through combination encoders to encoder selection and fusion.

\subsection{Problem Setup}
We detect answer correctness over a frozen language model $M$. In the
closed-book setting $M$ answers a question from parametric knowledge; in
the with-context setting retrieved passages precede the question. The two
settings map to a retrieval system's two decisions: whether to retrieve
and whether to trust the result.

A detector reads $M$'s internal signals along the single forward pass that
produced the answer and estimates $P(\text{answer correct})$; since $M$ is
frozen and nothing is regenerated, detection only reads
already-computed activations. The pass splits into context, question, and
answer spans, the structure the signals are encoded over
(Section~\ref{sec:signals}).

Labels come from judge consensus, separately per setting: three prompt
variants of an LLM judge score the setting's answer against the reference,
an example is kept only under unanimous agreement. The closed-book and with-context answers differ, so their
labels can differ, and each detector trains on its own.

\subsection{Internal Signals from a Single Forward Pass}
\label{sec:signals}
Let $M$ be a frozen transformer with $L$ layers and hidden size $d$, and
let the input be partitioned into the context, question, and answer spans.
At layer $\ell$ and position $t$ the residual stream carries a state
$h_t^{(\ell)}\in\mathbb{R}^{d}$, updated by the attention and feed-forward
contributions of that layer,
\begin{equation}
h_t^{(\ell)} = h_t^{(\ell-1)} + \mathrm{att}_t^{(\ell)} + \mathrm{ffn}_t^{(\ell)} .
\end{equation}
Writing $N$ for the final normalization and $W_U$ for the unembedding, the
next-token distribution at position $t$ is
$\pi_t = \operatorname{softmax}\!\big(W_U\,N(h_t^{(L)})\big)$, and $v_t$ is
the token the model emits. From this pass we form four token-level
feature families.

\paragraph{Hidden states.}
We take the residual state at a single layer $\ell^\star$,
\begin{equation}
\phi^{\mathrm{hid}}_t = h_t^{(\ell^\star)} \in \mathbb{R}^{d},
\end{equation}
the representation read by hidden-state probes of truthfulness
\citep{azaria2023internal,chen2024inside}; the layer $\ell^\star$ is chosen
on validation data by a sweep (Figure~\ref{fig:layer-sweep}).

\paragraph{Token probabilities.}
We reduce the next-token distribution to five scalars,
\begin{equation}
\begin{aligned}
\phi^{\mathrm{prob}}_t = \big(\,
& \pi_t(v_t),\; -\log \pi_t(v_t),\; \mathrm{H}(\pi_t),\\
& \max_w \pi_t(w),\; \pi_t^{(1)}-\pi_t^{(2)}\,\big),
\end{aligned}
\end{equation}
the emitted-token probability, its surprisal, the entropy $\mathrm{H}(\pi_t)$,
the top probability, and the margin between the two most likely tokens
($\pi_t^{(1)}\!\ge\!\pi_t^{(2)}$). These read a model's own probability as a
correctness signal \citep{kadavath2022,sriramanan2024llm}.

\paragraph{Residual-stream contributions.}
We decompose the pass by component. For each layer $\ell$
and each contribution $g\in\{\mathrm{att}_t^{(\ell)},\,\mathrm{ffn}_t^{(\ell)}\}$ added to the
running state $h$, we record its norm $\lVert g\rVert_2$ and its logit-lens
push toward the emitted token,
\begin{equation}
\begin{aligned}
\Delta_t(g) = {}& \big\langle W_U[v_t],\,N(h+g)-N(h)\big\rangle,\\
\phi^{\mathrm{res}}_t = {}& \big(\lVert g\rVert_2,\; \Delta_t(g)\big)_{\ell,\,g}
\in\mathbb{R}^{4L},
\end{aligned}
\end{equation}
where $W_U[v_t]$ is the unembedding row of $v_t$, so $\Delta_t(g)$ is how
much the component moves the emitted token's logit, and the two readings are
collected over every layer and both components. This reads the residual
stream in the spirit of mechanistic detectors
\citep{sun2025redeep,xiong2025toward}.

\paragraph{Attention distributions.}
Attention is causal and the spans are ordered, context before question
before answer, so between any two spans attention flows one way only: the
later span attends to the earlier one. To give every span a reading for each
of the other spans we therefore take each pairwise map in both directions.
Let $\alpha^{(\ell,i)}_{t\to s}$ be the weight from query $t$ to key $s$ at
layer $\ell$ and head $i$, with $t$ in the later span $V$ and $s$ in the
earlier span $U$, and let $\bar\alpha$ normalize each row over $U$ and
$\tilde\alpha$ each column over $V$. We record
\begin{equation}
\begin{aligned}
\phi^{\mathrm{out}}_{t} &= \Big(\mathrm{H}\big(\{\bar\alpha_{t\to s}\}_{s\in U}\big),\;
\textstyle\sum_{s\in U}\alpha_{t\to s}\Big),
&& t\in V,\\
\phi^{\mathrm{in}}_{s} &= \Big(\mathrm{H}\big(\{\tilde\alpha_{t\to s}\}_{t\in V}\big),\;
\textstyle\sum_{t\in V}\alpha_{t\to s}\Big),
&& s\in U,
\end{aligned}
\end{equation}
per layer and head, an entropy and a mass reading in each direction: the
entropy is low when the cross-span attention concentrates on a few tokens
and high when it is diffuse, and the mass is the share of $t$'s attention
budget spent on $U$ (rows of $\alpha$ sum to one) or, incoming, the total
weight $s$ receives from $V$, in the spirit of attention-based detectors
\citep{chuang2024lookback,binkowski2025hallucination}.

\subsection{Combination Encoders and Late Fusion}
\label{sec:selection}
The four families are not equally informative: a strong signal can take
over a jointly trained encoder, so weaker but complementary signals go
under-used. Encoding each signal separately and combining
only their predictions avoids this, but gives up any interaction between
signals at the token level, where some evidence appears only jointly. We
therefore build combination encoders. Each takes a subset of the signals and
stacks their features, at every token, into one vector, where the signals
meet; a token-level CNN encodes the context, question, and
answer spans, and an MLP over the three span vectors outputs the encoder's
estimate of $P(\text{answer correct})$. With four signals there are
$2^4-1=15$ subsets, hence $15$ encoders, all with the base model
frozen.

Only a few of them are worth keeping. We combine a selected set by late
fusion: a small trained head reads the chosen encoders' pooled
representations, and their predicted scores, into the final estimate. The set is chosen by greedy forward selection
on validation AUROC. The head and the stopping rule are detailed in
Appendix~\ref{app:arch}.

The candidate pool differs by setting. In the closed-book setting it is the
fifteen encoders trained on that pass. In the with-context setting the pool
also contains the closed-book encoders, giving thirty candidates. This is
almost free in a deployed system: the closed-book pass is run anyway to
decide whether to retrieve (Section~\ref{sec:pgir}), so caching its
per-token features makes those encoders available once a passage has been
retrieved. What they add is the model's judgement of the question before
the retrieved text arrived, which the with-context pass cannot produce.
 
\section{Experiments}
 
\subsection{Experimental Setup}

\paragraph{Data and labels.}
We evaluate on six open-domain QA datasets. Three multi-hop sets, MuSiQue
\citep{trivedi2022musique}, HotpotQA \citep{yang2018hotpotqa}, and
2WikiMultihopQA \citep{ho2020constructing}, form the in-domain data with a fixed
training/validation/test split; three single-hop sets, PopQA
\citep{mallen2023not}, Natural Questions (NQ) \citep{kwiatkowski2019natural}, and
TriviaQA \citep{joshi2017triviaqa}, are held out entirely for zero-shot
evaluation. Throughout we use a
fixed subsample of each dataset. The base model is
Qwen3.5-9B \citep{qwen2026qwen35} with greedy decoding. For each question the frozen model produces one answer, which a
Qwen3.5-27B judge labels correct or incorrect against the reference; we take the
unanimous vote of three prompt variants and build labels separately for each
setting, since the two settings elicit different answers; a blind human
re-annotation of 100 items agrees with the labels on $98\%$, Cohen's
$\kappa=0.96$ (splits, prompts, and label balance in Appendix~\ref{app:data}).

\paragraph{Settings and signals.}
The two settings are \emph{closed-book}, where the model answers from parametric
memory, and \emph{with-context}, where it is conditioned on retrieved passages.
The main tables use the datasets' own supporting passages as the context, so every
detector reads the same controlled input; Appendix~\ref{app:realrag} repeats the
comparison with passages from a real retriever, where the ranking carries over.
The with-context data augments each question with a rewritten passage, and every
trained detector, ours and the baselines, sees the same augmentation. All signals
are read from a single forward pass; the hidden-state signal is taken from
layer~17, selected on validation (Figure~\ref{fig:layer-sweep}). The encoder
architecture, layer sweep, and selection protocol are detailed in
Appendix~\ref{app:arch}; feature-construction details are in the supplementary material.

\paragraph{Metrics and protocol.}
The detector outputs $P(\text{answer correct})$; its complement scores errors. We
report AUROC, unchanged by this sign convention, and AUPRC with the incorrect
answer as the positive class. Every design choice, including encoder selection and
hyperparameters, is made on the validation split alone, and all numbers are three-seed
means; the greedy selection procedure is detailed in Appendix~\ref{app:arch}.

\paragraph{Baselines.}
We compare against 23 faithfully re-implemented detectors in three tiers:
sampling-based methods that draw ten extra generations per question, single-signal
single-pass probes, and detectors specific to retrieval-augmented generation. All
methods score the same greedy generations under the same labels.
 
\subsection{Detecting Answer Correctness Across Settings}

\begin{table*}[t]
\centering
\small
\setlength{\tabcolsep}{3pt}
\renewcommand{\arraystretch}{0.95}
\begin{tabular}{@{}l cccc cccc@{}}
\toprule
& \multicolumn{4}{c}{Closed-book} & \multicolumn{4}{c}{With context} \\
\cmidrule(lr){2-5} \cmidrule(lr){6-9}
& ID & \multicolumn{3}{c}{OOD} & ID & \multicolumn{3}{c}{OOD} \\
\cmidrule(lr){2-2} \cmidrule(lr){3-5} \cmidrule(lr){6-6} \cmidrule(lr){7-9}
Method & Test & PopQA & NQ & TQA & Test & PopQA & NQ & TQA \\
\midrule
\multicolumn{9}{l}{\emph{Sampling-based (10 extra generations per query)}} \\
Semantic Entropy~\citep{kuhn2023semantic} & .790/.891 & .876/.957 & .758/.751 & \underline{.917}/\underline{.808} & .773/.527 & .751/.811 & .561/.684 & .836/.567 \\
Bayesian SE~\citep{wang2023hallucination} & .801/.893 & .884/.957 & .758/.767 & \textbf{.921}/\textbf{.821} & .773/.522 & .753/.813 & .564/.692 & .840/.571 \\
SelfCheckGPT~\citep{manakul2023selfcheckgpt} & \underline{.805}/\underline{.898} & .881/.956 & \underline{.802}/\underline{.835} & .901/.779 & .699/.546 & .645/.809 & .511/.723 & .821/.581 \\
EigenScore~\citep{chen2024inside} & .689/.831 & .819/.927 & .762/.801 & .754/.554 & .861/.591 & .834/.861 & .660/.759 & .728/.401 \\
\midrule
\multicolumn{9}{l}{\emph{Single-signal, single forward pass}} \\
SAPLMA~\citep{azaria2023internal} & .795/.890 & \underline{.889}/\underline{.962} & .785/.827 & .843/.692 & \underline{.892}/\underline{.744} & \underline{.924}/\underline{.937} & \underline{.847}/\underline{.899} & \underline{.883}/\underline{.637} \\
LLMsKnow~\citep{ICLR2025_a712d461} & .795/.891 & .859/.947 & .749/.787 & .828/.646 & .884/.703 & .912/.935 & .831/.898 & .851/.620 \\
SEP~\citep{kossen2024semantic} & .761/.867 & .818/.928 & .639/.664 & .769/.542 & .789/.519 & .819/.864 & .657/.773 & .751/.458 \\
LLM-Check~\citep{sriramanan2024llm} & .747/.861 & .803/.938 & .686/.741 & .687/.557 & .779/.579 & .800/.875 & .759/.861 & .734/.499 \\
\midrule
\multicolumn{9}{l}{\emph{RAG-specific detectors (single forward pass)}} \\
Lookback-Lens~\citep{chuang2024lookback} & -- & -- & -- & -- & .851/.665 & .906/.932 & .777/.850 & .814/.532 \\
ReDeEP-PKS~\citep{sun2025redeep} & -- & -- & -- & -- & .664/.333 & .732/.736 & .666/.766 & .715/.378 \\
LUMINA~\citep{yeh2026lumina} & -- & -- & -- & -- & .596/.349 & .703/.803 & .629/.774 & .617/.327 \\
\midrule
\multicolumn{9}{l}{\emph{Multi-signal fusion (ours)}} \\
BEATRICE & \textbf{.839}/\textbf{.913} & \textbf{.924}/\textbf{.976} & \textbf{.814}/\textbf{.847} & .874/.762 & \textbf{.932}/\textbf{.826} & \textbf{.948}/\textbf{.965} & \textbf{.896}/\textbf{.941} & \textbf{.930}/\textbf{.774} \\
\bottomrule
\end{tabular}
\caption{Detecting incorrect answers of Qwen3.5-9B. Each cell is AUROC\,/\,AUPRC; per column and metric \textbf{bold} is best and \underline{underline} second, and ``--'' marks detectors that require context.}
\label{tab:main-detection}
\end{table*}

Table~\ref{tab:main-detection} compares BEATRICE with a representative subset of
the baselines, grouped by tier (full per-dataset panorama with AUPRC in
Appendix~\ref{sec:appendix-detection}). Across settings and datasets, BEATRICE is
the strongest detector on almost every column, from a single forward pass. The internal signals are what keep this
reliable out of distribution: text-only classifiers trained on the same data, up to a
fine-tuned transformer, come within a few points in-domain but trail BEATRICE on every
column and fall away sharply on the hardest out-of-distribution sets, the fine-tuned
encoder dropping below the linear models; Appendix~\ref{app:textonly} traces this to
surface correlations that do not transfer.

The only baselines that come close are the sampling-based consistency methods,
and on one column, closed-book TriviaQA, the best of them overtakes BEATRICE.
That is the regime the backbone knows best: it answers these single-hop popular-entity
questions correctly $71\%$ of the time closed-book, so agreement across resamples
tracks correctness closely there; once a passage is present, the same column goes to
BEATRICE. The edge is also narrow and costly: these
methods draw multiple generations per question, and on other datasets their agreement
signal falls toward chance, since a confidently and consistently wrong model can yield
samples that agree without being right.
A paired bootstrap against the strongest baseline on each column
(Appendix~\ref{app:significance}) finds the leads significant on every column with
context and in-domain closed-book, though not on the closed-book OOD columns.
On a second backbone Gemma\,4 \citep{gemmateam2026gemma4} BEATRICE again leads overall (full table
in Appendix~\ref{sec:appendix-gemma}).

\subsection{Anatomy of the Correctness Signal}
\label{sec:anatomy}

\begin{table*}[t]
\centering
\small
\setlength{\tabcolsep}{2pt}
\renewcommand{\arraystretch}{0.95}
\resizebox{\linewidth}{!}{
\begin{tabular}{@{}l c cccc c cccc@{}}
\toprule
& \multicolumn{5}{c}{Closed-book} & \multicolumn{5}{c}{With context} \\
\cmidrule(lr){2-6}\cmidrule(lr){7-11}
& & ID & \multicolumn{3}{c}{OOD} & & ID & \multicolumn{3}{c}{OOD} \\
\cmidrule(lr){3-3}\cmidrule(lr){4-6}\cmidrule(lr){8-8}\cmidrule(lr){9-11}
Variant & Encoder & Test & PopQA & NQ & TQA & Encoder & Test & PopQA & NQ & TQA \\
\midrule
  Hidden-only & \lp{}\gH{}\rp{} & .834/.910 & \underline{.917}/\underline{.973} & .795/.828 & .859/.735 & \lb{}\gH{}\rb{} & .929/.814 & .941/.957 & .880/.926 & .910/.738 \\
  w/o Hidden & \lp{}\gP{}\gR{}\gA{}\rp{}\lp{}\gP{}\rp{}\lp{}\gR{}\gA{}\rp{}\lp{}\gP{}\gA{}\rp{} & .832/.910 & .900/.969 & .754/.802 & .825/.685 & \lb{}\gP{}\gR{}\gA{}\rb{}\lb{}\gP{}\gA{}\rb{}\lp{}\gP{}\gR{}\gA{}\rp{} & .910/.780 & .937/.958 & .854/.916 & .891/.663 \\
  Early fusion & \lp{}\gH{}\gP{}\gR{}\gA{}\rp{} & \textbf{.839}/\textbf{.915} & .913/.972 & \underline{.813}/\underline{.841} & \underline{.870}/\underline{.755} & \lb{}\gH{}\gP{}\gR{}\gA{}\rb{} & .927/.808 & .944/.962 & .877/.926 & \underline{.922}/.739 \\
  Late fusion (single) & \lp{}\gH{}\rp{}\lp{}\gP{}\rp{}\lp{}\gR{}\rp{}\lp{}\gA{}\rp{} & \underline{.837}/.910 & .912/.971 & .785/.824 & .853/.721 & \lb{}\gH{}\rb{}\lb{}\gP{}\rb{}\lb{}\gR{}\rb{}\lb{}\gA{}\rb{} & .929/.819 & .942/.959 & .878/.928 & .917/.736 \\
  {\color{black!45} Late fusion (all)} & {\color{black!45} \lp{}\gS{}\rp{}$\times$15} & {\color{black!45} .841/.915} & {\color{black!45} .922/.974} & {\color{black!45} .815/.844} & {\color{black!45} .875/.759} & {\color{black!45} \lb{}\gS{}\rb{}$\times$15\;\lp{}\gS{}\rp{}$\times$15} & {\color{black!45} .931/.821} & {\color{black!45} .949/.967} & {\color{black!45} .889/.939} & {\color{black!45} .927/.766} \\
  w/o closed-book & — & —/— & —/— & —/— & —/— & \lb{}\gH{}\gP{}\gA{}\rb{}\lb{}\gH{}\gP{}\gR{}\rb{} & \underline{.931}/\underline{.824} & \underline{.946}/\underline{.963} & \underline{.890}/\underline{.933} & \textbf{.930}/\underline{.763} \\
  \textbf{Full (ours)} & \lp{}\gH{}\gR{}\rp{}\lp{}\gH{}\gP{}\gA{}\rp{}\lp{}\gH{}\gA{}\rp{} & \textbf{.839}/\underline{.913} & \textbf{.924}/\textbf{.976} & \textbf{.814}/\textbf{.847} & \textbf{.874}/\textbf{.762} & \lb{}\gH{}\gP{}\gA{}\rb{}\lb{}\gH{}\gP{}\gR{}\rb{}\lp{}\gH{}\gA{}\rp{} & \textbf{.932}/\textbf{.826} & \textbf{.948}/\textbf{.965} & \textbf{.896}/\textbf{.941} & \textbf{.930}/\textbf{.774} \\
\bottomrule
\end{tabular}
}
\caption{Feature and fusion ablation of BEATRICE. Each cell reports
\textbf{AUROC\,/\,AUPRC}, 3-seed mean.
The Encoder columns list the selected encoders in each setting: a bracket group is one encoder whose
glyphs are early-fused, and groups side by side are late-fused; \lb$\cdot$\rb\ marks a
with-context forward pass and \lp$\cdot$\rp\ a closed-book one. Glyphs: \gH{}~Hidden,
\gP{}~Prob, \gR{}~Resid, \gA{}~Attn. \textbf{Bold}~$=$~best,
\underline{underline}~$=$~second in each column per metric. The greyed \emph{Late fusion
(all)} row late-fuses every encoder; it is a saturation
reference, greedy selection matches it to within
noise, and it is excluded from the ranking.}
\label{tab:ablation}
\end{table*}

\begin{figure}[t]
\centering
\includegraphics[width=\columnwidth]{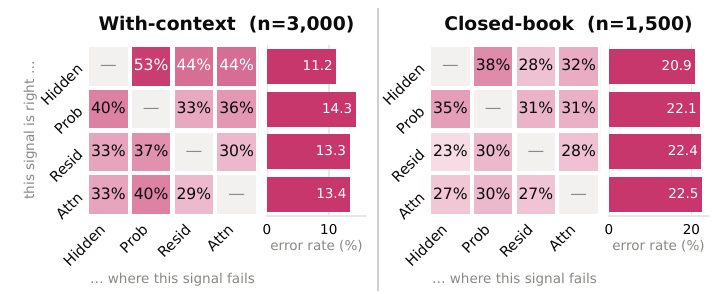}
\caption{Signal complementarity on the in-domain test set. Each cell is the fraction of the column signal's errors that the row signal classifies correctly, with every detector thresholded to flag its own true number of errors and columns normalised so the two settings compare; the bars give each signal's own error rate.}
\label{fig:complementarity}
\end{figure}

This is the core of the paper. Beyond assembling a detector, the search is a read-out:
what it keeps and what it discards is a map of where answer correctness can be read from
the model's internal state, predictive rather than mechanistic: which signals carry usable
evidence, not where the model computes it. Importantly, predicting final-answer correctness does not certify the validity of the underlying reasoning process: multi-hop QA systems can produce correct final answers despite failing intermediate sub-questions \cite{tang-etal-2021-multi}. Our analysis is therefore a predictive read-out of answer correctness. We read the map off the selection in both
settings. The four signals are
far from equally strong: read on its own, the hidden-state signal is by a wide margin the
most accurate, and the token-probability, residual-stream, and attention signals
each trail it. The strong signal is nonetheless not self-sufficient.
Figure~\ref{fig:complementarity} counts, among the errors each single-signal
detector makes, how many another signal gets right: the weaker signals recover a
third to a half of the hidden-state signal's mistakes, and no pairing is fully
redundant. The signals are complementary, so a detector drawing on all of
them has headroom the hidden-state signal alone does not. The paired bootstrap of
Appendix~\ref{app:significance} locates that headroom: the fused champion improves on the
hidden-state encoder in every column, significantly so on the held-out single-hop
sets ($+0.016$ to $+0.021$ AUROC), where a single signal is most brittle.

\begin{figure}[t]
\centering
\includegraphics[width=\columnwidth]{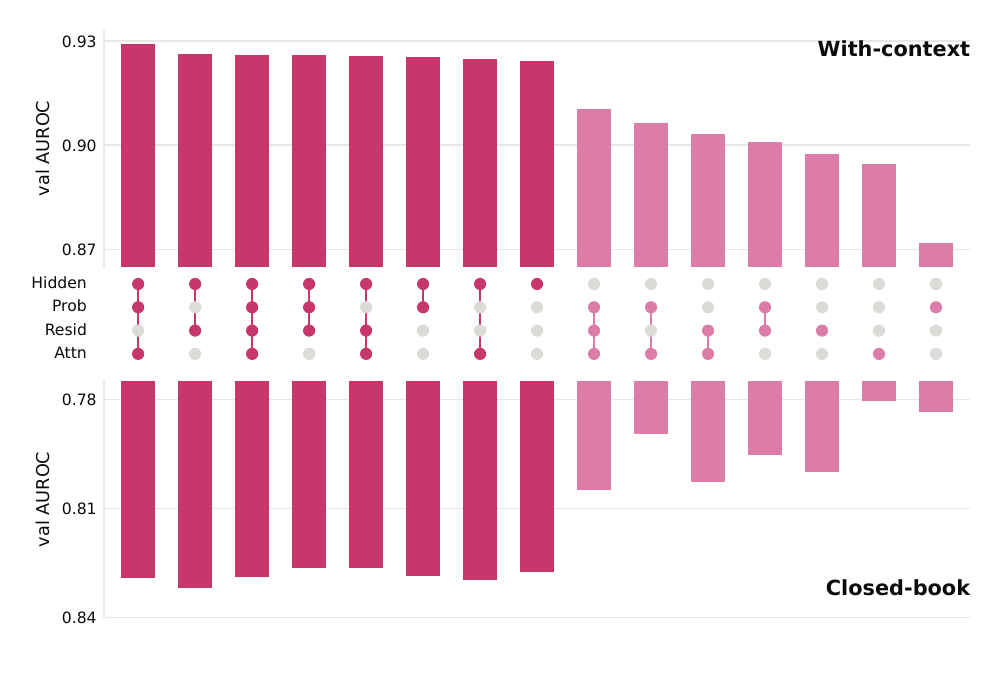}
\caption{Validation AUROC of all fifteen combination encoders, ordered best to worst. The dot matrix marks the signals each encoder fuses, and bars are darker when the encoder includes the hidden-state signal. AUROC axes are cropped to the observed range.}
\label{fig:signal-synergy}
\end{figure}

That headroom is not captured by merging everything at once.
Figure~\ref{fig:signal-synergy} scores all fifteen combination encoders, one per
subset of the four signals: encoders that include the hidden-state signal fill the
top of the ranking in both settings, yet the best encoder fuses three signals
rather than four, and adding the last one does not help. Table~\ref{tab:ablation}
tells the same story from the fusion side: a single encoder packing all four signals
trails the selected combination, while late-fusing every encoder only matches it, at more
cost. Selection reaches that ceiling with a few encoders, recovering the weaker,
complementary signals that an all-in-one encoder under-uses.

\begin{figure}[t]
\centering
\includegraphics[width=\columnwidth]{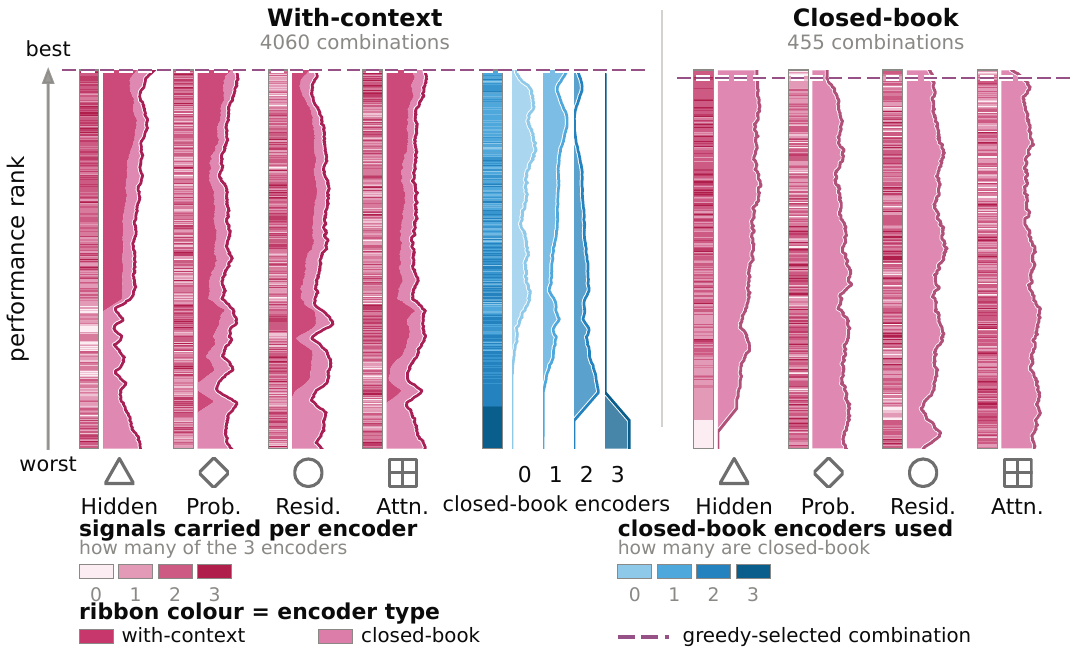}
\caption{Exhaustive ranking of all three-encoder combinations, best to worst. The ribbon is the local mean of how many of the three encoders carry that signal, split by whether the carrying encoder is with-context (dark) or closed-book (light); for the with-context pool the four wedges track how the closed-book count concentrates along the ranking. The dashed line marks the selected combination.}
\label{fig:combination-search}
\end{figure}

The selection also reaches across the two settings. Extending the candidate pool to include the closed-book encoders alongside the with-context ones, the greedy search keeps one closed-book encoder in the with-context detector. Figure~\ref{fig:combination-search} enumerates all three-encoder combinations and confirms two things. First, the greedy search is sound: in both settings the combination it selects ranks at or near the very top of the exhaustive ordering. Second, the cross-setting pattern is not an artifact of the greedy path: reading the closed-book-count panel toward the top, the best combinations concentrate on exactly one closed-book encoder, the ten best all carry one, while the worst are built entirely from them. On this backbone, then, the model's judgement before it sees the passage remains a useful extra viewpoint once context is present. The pattern is base-model dependent: on Gemma\,4 the best combinations shed the closed-book encoder at the top of the ranking (Appendix~\ref{sec:appendix-gemma}).

Appendix~\ref{app:anatomy-signals} takes this anatomy inside the trained
encoders. Ablating one signal family at a time shows that hidden states set
almost every decision, while each auxiliary family corrects a small,
setting-specific slice of the rest and acts only where the encoder is still
undecided; the one pair of signals whose joint encoder beats both of its
members, on either backbone, is token probabilities with attention; and what
the attention family responds to, verbatim support in the passage, ties its
reliability to retrieval quality.

Finally, the read-out draws almost entirely on the answer span. Zeroing each token region in turn and retraining, removing the answer tokens costs the most by far, about 0.02 AUROC with context and 0.01 closed-book, whereas removing the question or the retrieved context costs little, at most 0.006. The detector reads the model's handling of the context off the answer tokens rather than the context tokens themselves; the per-span breakdown is in Appendix~\ref{app:span} (Table~\ref{tab:span-ablation}).

\section{Probe-Gated Iterative Retrieval}
\label{sec:pgir}

\subsection{Controller: Pre- and Post-Retrieval Decisions}

\begin{figure}[t]
\centering
\includegraphics[width=\columnwidth]{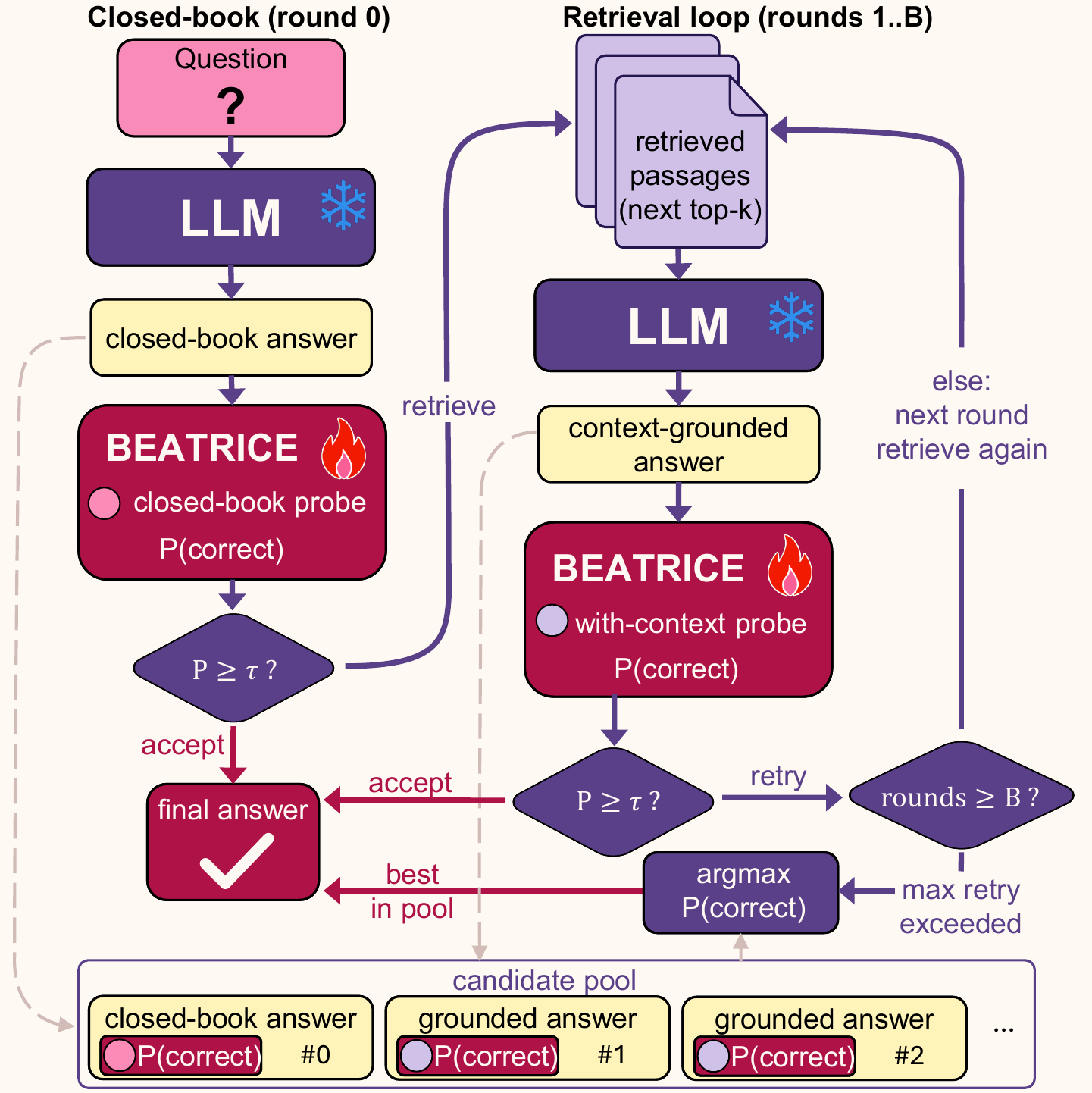}
\caption{PGIR control flow. A draft is accepted when its calibrated correctness
probability clears $\tau$; otherwise the controller retrieves and re-answers, up to a
budget of $B$ rounds, then returns the highest-scoring draft in the pool.}
\label{fig:pgir-flow}
\end{figure}

Because BEATRICE scores an answer the model has already produced, it doubles as a
retrieval controller: one score can decide whether a draft is trustworthy before a
retrieval is spent, and which draft to keep after one. Figure~\ref{fig:pgir-flow} shows
the control flow. The model first answers closed-book; if the closed-book probe clears an
acceptance threshold $\tau$ the draft is returned with no retrieval. Otherwise PGIR
retrieves and re-answers with context, and the with-context probe scores the grounded
draft; retrieval repeats up to a budget of $B$ rounds, returning the first draft to clear
$\tau$ or, failing that, the highest-scoring draft in the pool. Retrieval uses
E5~\citep{wang2022e5} over Wikipedia-2018 with $k{=}3$. Each round issues a new query, the
question augmented by the current draft's answer, and keeps only passages not retrieved in
earlier rounds, so every round adds fresh evidence; when that answer-guided query stops
turning up new passages the retriever pages deeper into the ranking instead. Since both
probes read passes the model already runs, the controller adds no generations of its own.

The score each probe reports is a calibrated correctness probability. On the validation
split we fit one isotonic regression per setting, mapping raw probe outputs to answer
correctness, so the acceptance test compares a genuine probability against $\tau$ rather
than an uncalibrated score. The threshold $\tau$ and the budget $B$ are chosen together
by a small grid search on the development split, taking the configuration with the
highest exact-match under a cap on the average number of generations per question; this
selects $\tau{=}0.6$ and $B{=}4$, used unchanged at test time. The test sets take no
part in the calibration or this selection.

\subsection{End-Task Results}

Table~\ref{tab:pgir-application} (Appendix~\ref{app:pgir-results}) reports end-task answer
quality across the confidence-control family, every system sharing the one backbone and retriever.
PGIR leads on macro exact-match and macro F1, and on the exact-match of four of the
six datasets, ahead of the adaptive baselines. This is the point of the demonstration: the
same read-out, unchanged, improves the answers themselves and not only our ability to audit
them.

\begin{figure}[t]
\centering
\includegraphics[width=\columnwidth]{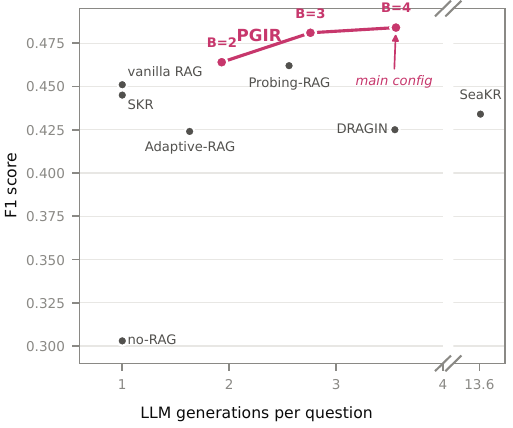}
\caption{Efficiency--accuracy frontier: macro F1 against generations per question, a
serving-cost proxy (the $x$-axis is broken for the far-right system). Points sweep
PGIR's budget $B$; $B{=}4$ is the reported configuration.}
\label{fig:pgir-pareto}
\end{figure}

The gain comes from where the signal is read. Systems that decide before retrieving,
whether from a learned router or a closed-book confidence estimate, stay close to the
vanilla single-retrieval baseline; systems that trigger further retrieval from
token-level or sampling uncertainty do not close the gap either, and the strongest of
them pays for its accuracy with over thirteen generations per question. PGIR reads the
signal after retrieval, scoring each grounded draft, and reaches the best accuracy at
fewer than four generations per question. Figure~\ref{fig:pgir-pareto} shows the
resulting frontier: sweeping the budget traces a curve that stays above the other
systems at every cost. Appendix~\ref{app:cases} walks through individual questions where a
low closed-book score triggers retrieval and a high with-context score earns trust.

The controller also transfers without retraining. The probe is never fit on PopQA, NQ,
or TriviaQA, yet PGIR leads on the first two of these out-of-distribution sets. Two
limits are worth stating plainly. On the two hardest multi-hop sets a far more expensive
loop overtakes it, buying its accuracy with the thirteen-generation budget just noted.
On single-hop TriviaQA a single retrieval already suffices, so the room for a verifier
to add F1 is small and vanilla retrieval edges ahead on that one metric, though PGIR
still leads its exact-match. On a second backbone (Gemma\,4) the controller transfers
just as well, with PGIR again first on macro exact-match and F1; the full table is in
Appendix~\ref{sec:appendix-gemma}.

\section{Conclusion}

Reading four families of internal signal from a single forward pass, and searching over
which to combine, we map where answer correctness can be read out of a frozen model: one family
dominates, the rest are complementary rather than redundant, the cue concentrates in the
answer span, and whether the model's judgement before it sees a passage stays informative
afterwards depends on the base model. The closed-book and with-context settings share this
anatomy in broad strokes but not in detail, so we read it separately in each. The
same read-out, without retraining, is accurate in and out of domain, and it can gate a
retrieval-augmented system as one downstream use. What we expect to carry is the anatomy
rather than any single score: knowing which families hold evidence, where it
concentrates, and when a further family adds information turns probe design from trial
and error over encoders into a measurement of the base model, one that can be repeated
as models change.

\bibliographystyle{plainnat}
\bibliography{main}

\clearpage
\appendix

\section{Datasets, Labels, and Prompts}
\label{app:data}
\paragraph{Datasets and splits.}
We draw on six open-domain QA datasets. The three multi-hop sets, MuSiQue, HotpotQA, and
2WikiMultihopQA, form the in-domain data: we sample $10{,}000$ training examples from each,
$30{,}000$ in all, and split each dataset's public development set evenly into validation
and test, giving $1{,}500$ validation and $1{,}500$ test ($500$ per dataset in each). This
split is fixed once and reused throughout. The three single-hop sets are held out entirely
for zero-shot evaluation: PopQA ($1{,}000$), NQ ($1{,}182$), and TriviaQA ($1{,}277$).
Every split is fixed by example id, so no example moves between training and evaluation
across settings or baselines.

\paragraph{Context and augmentation.}
In the with-context setting the passages are the datasets' own supporting paragraphs, not
retriever output; retrieval enters only in the downstream controller
(Section~\ref{sec:pgir}). We double the with-context data with a faithful normalization
rewrite, cast by the same larger model that assigns the labels (Qwen3.5-27B, greedy): it
turns each passage into a compact,
question-aligned evidence passage, preserving every entity, date, number, and relation,
using only information already present, and adding no facts or answers (the rewrite
prompts appear with the others below). The backbone re-answers on the rewritten pair and it
is labeled the same way, so each with-context example appears in an original and a
normalized form, and the with-context training and evaluation sets are twice the counts
above. Every trained detector, ours and the baselines, sees this same augmentation, so it
does not affect the comparison; it is a generic robustness measure unrelated to the
correctness task, and the controller of Section~\ref{sec:pgir} does not use it, retrieving
with E5 instead.

\paragraph{Labels.}
For each question the model produces an answer, which is judged correct or incorrect
against the reference answers. The label is the unanimous vote of the three judge prompts
below, cast by a larger model of the same family; we keep only the
examples on which all three rubrics agree, rather than break ties by majority.
Disagreement is rare: this filter discards $2.3\%$ of examples on Qwen3.5-9B. Labels are
built separately for the closed-book and with-context settings, since the two elicit
different answers. As a check on the judge, we re-annotated by hand 100 evaluation items
sampled uniformly at random (50 per setting), blind to the judge verdicts: human and
judge labels agree on $98\%$ of them, Cohen's $\kappa=0.96$ (bootstrap $95\%$ CI
$[0.90, 1.00]$).
Table~\ref{tab:label-dist} reports the resulting class balance for
every split under both settings.

\begin{table}[t]
\centering
\small
\setlength{\tabcolsep}{5pt}
\resizebox{\columnwidth}{!}{%
\begin{tabular}{ll r c cc}
\toprule
& & & Closed-book & \multicolumn{2}{c}{With-context} \\
\cmidrule(lr){4-4}\cmidrule(lr){5-6}
Split & Dataset & $N$ & Correct (\%) & Orig. & Rewr. \\
\midrule
\multirow{4}{*}{Train}
 & HotpotQA        & $10{,}000\,{\times}2$ & 39.8 & 95.2 & 95.5 \\
 & MuSiQue         & $10{,}000\,{\times}2$ & 17.8 & 60.8 & 62.5 \\
 & 2WikiMultihopQA & $10{,}000\,{\times}2$ & 42.8 & 95.4 & 96.0 \\
 & \emph{All}      & $30{,}000\,{\times}2$ & 33.5 & 83.8 & 84.7 \\
\midrule
\multirow{4}{*}{Val}
 & HotpotQA        & $500\,{\times}2$  & 40.4 & 92.0 & 91.6 \\
 & MuSiQue         & $500\,{\times}2$  & 14.8 & 56.4 & 58.8 \\
 & 2WikiMultihopQA & $500\,{\times}2$  & 34.6 & 85.8 & 87.6 \\
 & \emph{All}      & $1{,}500\,{\times}2$ & 29.9 & 78.1 & 79.3 \\
\midrule
\multirow{4}{*}{Test}
 & HotpotQA        & $500\,{\times}2$  & 37.8 & 93.4 & 93.4 \\
 & MuSiQue         & $500\,{\times}2$  & 14.0 & 51.4 & 53.0 \\
 & 2WikiMultihopQA & $500\,{\times}2$  & 36.4 & 89.4 & 90.0 \\
 & \emph{All}      & $1{,}500\,{\times}2$ & 29.4 & 78.1 & 78.8 \\
\midrule
\multirow{3}{*}{OOD}
 & PopQA    & $1{,}000$ & 22.4 & 39.3 & --   \\
 & NQ       & $1{,}182$ & 42.7 & 32.5 & --   \\
 & TriviaQA & $1{,}277$ & 70.8 & 78.1 & --   \\
\bottomrule
\end{tabular}%
}
\caption{Correctness-label distribution across all splits, reported as the fraction of
examples the backbone answers correctly (the positive class; the incorrect fraction is the
complement). $N$ is the number of examples in that split; for with-context we give the rate
on the original passages (\emph{Orig.}) and on the normalization rewrite (\emph{Rewr.}) of
Appendix~\ref{app:data} separately; ${\times}2$ marks the splits whose with-context set
combines both and is thus twice $N$.}
\label{tab:label-dist}
\end{table}

\paragraph{Prompts.}
The prompts appear below. The answering prompt asks for a direct final answer and is the
input whose single forward pass the probe reads; the three judge prompts share a format
and differ only in the rubric they apply; and the two rewrite prompts produce the
with-context augmentation. In each, \ph{angled} text marks a filled-in placeholder.

\promptcard{Answering prompt (read for feature extraction)}{%
\ph{retrieved passage, with-context only}\\[2pt]
Please answer the following question based on your knowledge\ph{ and the context}.\\[2pt]
Question: \ph{question}\\[2pt]
Directly answer with the final answer without any explanation or reasoning process:%
}

\promptcard{Judge 1 $\cdot$ strict semantic}{%
You are a strict but fair semantic answer judge.\\[2pt]
Question: \ph{question}\quad Gold answer(s): \ph{gold}\quad Model answer: \ph{answer}\\[2pt]
Rules: judge only the final answer; accept aliases, abbreviations, spelling variants, and equivalent entity names; mark partial answers incorrect when a more specific answer is asked; accept a coarser date only at the granularity the question asks; ``I don't know'', empty, or refusal is Incorrect.\\[2pt]
Return exactly one token: Correct or Incorrect.%
}

\promptcard{Judge 2 $\cdot$ alias-friendly}{%
Decide whether the model answer is acceptable for the question.\\[2pt]
Question, acceptable gold answers, model answer: \ph{\dots}\\[2pt]
Use an alias-friendly standard: Correct if it refers to the same entity or value as any gold answer; Correct if it is a common short name, stage name, abbreviation, or harmless formatting variant; Incorrect if it names a different entity, relation, time, place, or number, or gives only a vague category.\\[2pt]
Return exactly: Correct or Incorrect.%
}

\promptcard{Judge 3 $\cdot$ decomposed fact}{%
Check whether the model answer satisfies the information need.\\[2pt]
Question, gold answer(s), model answer: \ph{\dots}\\[2pt]
Procedure: identify the entity, relation, or type of answer the question requests; compare the model answer to the gold answer(s); ignore extra wording if the final answer is unambiguous.\\[2pt]
Output exactly one word: Correct or Incorrect.%
}

\promptcard{Rewrite $\cdot$ context (with-context augmentation)}{%
Rewrite the context into a compact evidence-style passage aligned to the question.\\[2pt]
Rules: use only information from the context; preserve named entities, dates, numbers, and relations that may help answer the question; remove irrelevant wording and formatting noise; do not answer the question directly unless the context states the answer.\\[2pt]
Output: Normalized Context: \ph{rewritten context}\\[2pt]
Question: \ph{question}\quad Context: \ph{context}%
}

\section{Model Architecture and Training}
\label{app:arch}
This appendix details the combination-encoder architecture sketched in the main text and
the protocol used to select and fuse encoders.

\paragraph{Per-encoder architecture.}
A combination encoder reads a subset of the four signal families. At every token the
selected families are projected to a common width and stacked into one vector, which is
where the signals interact. Each of the context, question, and answer spans is then
encoded by its own token-level residual convolutional network: a width-1 convolution
projects the stacked features to a hidden width of $128$, followed by three residual
blocks, each two convolutions of kernel size $3$ with GroupNorm and GELU and dropout, all
masked so padding never contributes. Each span is reduced to a vector by concatenating a
masked max-pool and a masked mean-pool over its tokens. The three span vectors and their
pairwise differences are concatenated and passed through a two-layer MLP with LayerNorm
and GELU, whose output is the encoder's estimate of $P(\text{answer correct})$; the
closed-book encoder uses the question and answer spans only. The base model, its token
embeddings, and the hidden states are frozen throughout; the only regularization is a
small Gaussian noise added to the input features during training together with the
dropout above.

\paragraph{Late fusion.}
The selected encoders are frozen. The fusion head takes, over the selected encoders, each
pooled representation (of width $2\times128$), each predicted score, and the absolute
pairwise differences of those scores, concatenates them, and maps the result through a
two-layer MLP with LayerNorm and GELU to the final logit. Reading the score disagreements
lets the head weight an encoder by whether the others corroborate it. During training each
encoder is dropped independently with probability $0.3$, so the head cannot come to rely on
any single one. Scores are turned into calibrated probabilities as described in the
controller section.

\begin{figure}[t]
\centering
\includegraphics[width=\columnwidth]{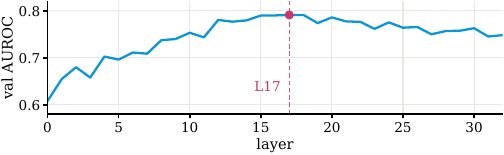}
\caption{Validation layer sweep: correctness AUROC of the hidden-state signal read from
each layer. Layer~17 is the maximum and is used throughout.}
\label{fig:layer-sweep}
\end{figure}

\paragraph{Layer selection.}
The hidden-state family is read from a single layer chosen on the validation split by the
sweep in Figure~\ref{fig:layer-sweep}; layer~$17$ is the maximum and is used throughout.

\paragraph{Selection protocol.}
Encoders are chosen by greedy forward selection on seed-averaged validation AUROC: starting
from the empty set, each round trains a fresh fusion head for every not-yet-selected
encoder added to the current set, keeps the one that most improves the fused score, and
stops when the best remaining addition helps by less than $5\times10^{-4}$
(Algorithm~\ref{alg:selection}). All selection uses the validation split only, and every
reported number is the mean over three training seeds.

\begin{algorithm}[t]
\caption{Greedy encoder selection with late fusion}
\label{alg:selection}
\begin{algorithmic}[1]
\REQUIRE frozen candidate encoders $\mathcal{E}$; validation set $V$; margin $\delta{=}5\times10^{-4}$
\STATE $S \leftarrow \varnothing$;\quad $a^{\star} \leftarrow 0.5$
\WHILE{$\mathcal{E}\setminus S \neq \varnothing$}
  \FOR{$e \in \mathcal{E}\setminus S$}
    \STATE train a fusion head on $S\cup\{e\}$;\; \\
    $a_e \leftarrow$ seed-averaged $\mathrm{AUROC}$ on $V$
  \ENDFOR
  \STATE $e^{\star} \leftarrow \arg\max_{e}\, a_e$
  \IF{$a_{e^{\star}} - a^{\star} \le \delta$}
    \STATE \textbf{break}
  \ENDIF
  \STATE $S \leftarrow S\cup\{e^{\star}\}$;\quad $a^{\star} \leftarrow a_{e^{\star}}$
\ENDWHILE
\STATE retrain the fusion head on $S$
\RETURN selected set $S$ and its fusion head
\end{algorithmic}
\end{algorithm}

\paragraph{Training.}
Each encoder and each fusion head is trained with AdamW (learning rate $2\times10^{-4}$,
weight decay $0.01$) for up to $60$ epochs at batch size $512$, halving the learning rate
after four epochs without a validation gain and stopping early after ten. The base model,
its token embeddings, and the hidden states stay frozen throughout. All experiments ran
on a Linux server with NVIDIA H100 80GB GPUs, using PyTorch and Transformers for
feature extraction and training and a vLLM OpenAI-compatible server for generation.

\section{Full Per-Dataset Detection Results}
\label{sec:appendix-detection}
Table~\ref{tab:main-detection} in the main text reports a representative subset of
detectors on the aggregate in-domain test split and the OOD splits. For completeness,
Tables~\ref{tab:appendix-detection} (closed-book) and~\ref{tab:appendix-detection-ctx}
(with context) give the full panorama: BEATRICE and all 23 faithfully re-implemented
baselines, broken out by individual in-domain dataset (MuSiQue, HotpotQA,
2WikiMultihopQA) as well as the aggregate, and reporting both AUROC and AUPRC (with the
incorrect answer as the positive class, so the random-guess AUPRC equals the per-column
error rate).
\begin{table*}[t]
\centering
\footnotesize
\setlength{\tabcolsep}{4pt}
\begin{tabular}{l cccc ccc}
\toprule
& \multicolumn{4}{c}{In-domain} & \multicolumn{3}{c}{Out-of-distribution} \\
\cmidrule(lr){2-5} \cmidrule(lr){6-8}
Method & MuS & HoP & 2Wi & Test & PopQA & NQ & TQA \\
\midrule
\emph{Random AUPRC (error rate)} & \emph{.860} & \emph{.622} & \emph{.636} & \emph{.706} & \emph{.776} & \emph{.573} & \emph{.292} \\
\midrule
BEATRICE & \textbf{.804}/\textbf{.955} & .844/.866 & \textbf{.792}/\textbf{.871} & \textbf{.839}/\textbf{.913} & \textbf{.924}/\textbf{.976} & \textbf{.814}/\textbf{.847} & .874/.762 \\
SelfCheckGPT-NLI & .756/.944 & \textbf{.860}/\textbf{.909} & .789/.855 & .805/.898 & .881/.956 & .802/.835 & .901/.779 \\
Bayesian SE & .708/.924 & .848/.883 & .776/.862 & .801/.893 & .884/.957 & .758/.767 & \textbf{.921}/\textbf{.821} \\
LLMsKnow & .735/.934 & .802/.848 & .757/.853 & .795/.891 & .859/.947 & .749/.787 & .828/.646 \\
SAPLMA (mean) & .758/.935 & .778/.836 & .768/.866 & .795/.890 & .889/.962 & .785/.827 & .843/.692 \\
Semantic Entropy & .685/.921 & .839/.884 & .755/.853 & .790/.891 & .876/.957 & .758/.751 & .917/.808 \\
Semantic Entropy (discrete) & .683/.915 & .840/.879 & .754/.847 & .790/.885 & .878/.955 & .760/.770 & .916/.805 \\
SAPLMA (last) & .729/.930 & .777/.812 & .763/.864 & .787/.882 & .860/.945 & .756/.792 & .814/.645 \\
SelfCheckGPT-ngram & .705/.924 & .833/.882 & .670/.754 & .764/.872 & .906/.966 & .787/.840 & .874/.755 \\
SEP & .696/.926 & .766/.805 & .736/.837 & .761/.867 & .818/.928 & .639/.664 & .769/.542 \\
LLM-Check (combined) & .629/.907 & .766/.815 & .719/.831 & .747/.861 & .803/.938 & .686/.741 & .687/.557 \\
LLM-Check (entropy) & .620/.917 & .721/.795 & .698/.825 & .716/.857 & .786/.925 & .651/.712 & .657/.517 \\
EigenScore & .584/.896 & .754/.811 & .670/.784 & .689/.831 & .819/.927 & .762/.801 & .754/.554 \\
LLM-Check (hidden) & .590/.898 & .582/.690 & .547/.676 & .568/.756 & .599/.825 & .531/.596 & .548/.326 \\
LLM-Check (perplexity) & .583/.902 & .495/.622 & .455/.616 & .501/.712 & .567/.791 & .496/.571 & .465/.265 \\
LLM-Check (attention) & .600/.911 & .502/.582 & .478/.589 & .496/.663 & .618/.807 & .573/.636 & .588/.374 \\
\bottomrule
\end{tabular}
\caption{Closed-book detection panorama: AUROC\,/\,AUPRC on every
dataset for all baselines and BEATRICE. Positive class for AUPRC is the
incorrect answer; the italic row is the random-guess AUPRC (per-column error rate).
Bold marks the best in each column across all rows. In-domain test sizes: MuSiQue 499,
HotpotQA and 2WikiMultihopQA 500 each; OOD covers PopQA/NQ/TriviaQA on which the probe
is zero-shot.}
\label{tab:appendix-detection}
\end{table*}

\begin{table*}[t]
\centering
\footnotesize
\setlength{\tabcolsep}{4pt}
\begin{tabular}{l cccc ccc}
\toprule
& \multicolumn{4}{c}{In-domain} & \multicolumn{3}{c}{Out-of-distribution} \\
\cmidrule(lr){2-5} \cmidrule(lr){6-8}
Method & MuS & HoP & 2Wi & Test & PopQA & NQ & TQA \\
\midrule
\emph{Random AUPRC (error rate)} & \emph{.486} & \emph{.066} & \emph{.106} & \emph{.219} & \emph{.607} & \emph{.675} & \emph{.219} \\
\midrule
BEATRICE & \textbf{.874}/\textbf{.869} & \textbf{.829}/\textbf{.437} & \textbf{.954}/\textbf{.800} & \textbf{.932}/\textbf{.826} & \textbf{.948}/\textbf{.965} & \textbf{.896}/\textbf{.941} & \textbf{.930}/\textbf{.774} \\
SAPLMA (mean) & .819/.822 & .726/.209 & .925/.738 & .892/.744 & .924/.937 & .847/.899 & .883/.637 \\
LLMsKnow & .815/.781 & .743/.250 & .929/.671 & .884/.703 & .912/.935 & .831/.898 & .851/.620 \\
SAPLMA (last) & .794/.775 & .760/.269 & .924/.626 & .883/.696 & .908/.930 & .805/.875 & .856/.601 \\
EigenScore & .747/.683 & .687/.207 & .892/.396 & .861/.591 & .834/.861 & .660/.759 & .728/.401 \\
Lookback-Lens & .755/.760 & .717/.247 & .886/.477 & .851/.665 & .906/.932 & .777/.850 & .814/.532 \\
SelfCheckGPT-ngram & .755/.736 & .798/.231 & .729/.275 & .817/.580 & .874/.900 & .702/.817 & .895/.700 \\
SEP & .653/.626 & .621/.176 & .798/.412 & .789/.519 & .819/.864 & .657/.773 & .751/.458 \\
LLM-Check (combined) & .718/.724 & .622/.132 & .713/.345 & .779/.579 & .800/.875 & .759/.861 & .734/.499 \\
Bayesian SE & .688/.658 & .731/.246 & .731/.295 & .773/.522 & .753/.813 & .564/.692 & .840/.571 \\
Semantic Entropy (discrete) & .686/.652 & .711/.257 & .741/.284 & .773/.520 & .751/.808 & .565/.690 & .838/.565 \\
Semantic Entropy & .686/.658 & .713/.278 & .738/.286 & .773/.527 & .751/.811 & .561/.684 & .836/.567 \\
SelfCheckGPT-NLI & .658/.683 & .652/.292 & .610/.343 & .699/.546 & .645/.809 & .511/.723 & .821/.581 \\
ReDeEP-PKS & .592/.531 & .616/.139 & .538/.140 & .664/.333 & .732/.736 & .666/.766 & .715/.378 \\
LLM-Check (attention) & .651/.626 & .461/.060 & .737/.189 & .663/.302 & .594/.630 & .595/.708 & .618/.266 \\
ReDeEP & .586/.530 & .608/.123 & .528/.138 & .658/.331 & .734/.744 & .668/.768 & .716/.389 \\
LLM-Check (entropy) & .655/.682 & .557/.121 & .515/.222 & .655/.461 & .783/.870 & .697/.835 & .662/.433 \\
LLM-Check (perplexity) & .596/.613 & .503/.070 & .793/.353 & .621/.359 & .549/.676 & .538/.691 & .598/.280 \\
LUMINA & .650/.623 & .445/.066 & .456/.112 & .596/.349 & .703/.803 & .629/.774 & .617/.327 \\
LUMINA-IPR & .559/.579 & .537/.075 & .498/.126 & .594/.340 & .671/.782 & .611/.763 & .584/.305 \\
LUMINA-MMD & .704/.694 & .429/.059 & .486/.114 & .536/.224 & .556/.601 & .566/.687 & .570/.256 \\
ReDeEP-ECS & .498/.515 & .518/.077 & .465/.097 & .536/.278 & .682/.785 & .577/.740 & .661/.413 \\
LLM-Check (hidden) & .479/.545 & .583/.151 & .295/.114 & .470/.291 & .577/.726 & .619/.779 & .545/.275 \\
\bottomrule
\end{tabular}
\caption{Retrieval-augmented (with-context) detection panorama: AUROC\,/\,AUPRC on every
dataset for all baselines and BEATRICE, including the RAG-specific detectors
(Lookback-Lens, ReDeEP variants, LUMINA variants). Conventions as in
Table~\ref{tab:appendix-detection}. In-domain test sizes: 500 per multi-hop set.}
\label{tab:appendix-detection-ctx}
\end{table*}

\section{Statistical Significance of the Detection Gains}
\label{app:significance}
To check that the leads in Table~\ref{tab:main-detection} are not sampling artifacts, we
run a paired bootstrap on each column against the strongest of all $23$ baselines there,
taken over the full panorama of Appendix~\ref{sec:appendix-detection} rather than the
representative subset of Table~\ref{tab:main-detection}. Both detectors score the same
examples, so we resample examples ($10^4$ draws) and read the $95\%$ interval and
two-sided $p$ of the AUROC difference directly; this accounts for the correlation
between the two scores, which separate per-method intervals would not.
Table~\ref{tab:significance} reports the result. The lead is significant on all four
with-context columns and on the in-domain closed-book column ($p\le.007$; all five
survive a Holm correction across the eight tests). The three closed-book OOD columns are
not significant: on PopQA the champion beats every single-pass probe decisively but its
margin over the best sampling method does not clear the interval ($+.018$, $p=.09$), on
NQ the margin is small ($+.012$), and on TriviaQA the sampling methods lead outright
($-.046$), the regime discussed in Appendix~\ref{app:textonly} where the model already
knows the answer and its own consistency suffices. With retrieved context, the setting in
which a correctness detector is most useful, every lead is significant.

\begin{table}[t]
\centering
\small
\setlength{\tabcolsep}{3pt}
\resizebox{\columnwidth}{!}{%
\begin{tabular}{ll l c c}
\toprule
Setting & Data & Strongest baseline & $\Delta$AUROC [$95\%$ CI] & $p$ \\
\midrule
\multirow{4}{*}{\shortstack[l]{Closed-\\book}}
 & Test  & SelfCheck-NLI (.805)   & $+.034$ $[+.010,+.060]^\dagger$ & $.007$ \\
 & PopQA & SelfCheck-ngram (.906) & $+.018$ $[-.003,+.037]$         & $.091$ \\
 & NQ    & SelfCheck-NLI (.802)   & $+.012$ $[-.016,+.041]$         & $.41$ \\
 & TQA   & Bayesian SE (.921)     & $-.046$ $[-.066,-.026]^\dagger$ & $<\!10^{-4}$ \\
\midrule
\multirow{4}{*}{\shortstack[l]{With\\context}}
 & Test  & SAPLMA (.892)          & $+.034$ $[+.021,+.049]^\dagger$ & $<\!10^{-4}$ \\
 & PopQA & SAPLMA (.924)          & $+.025$ $[+.014,+.036]^\dagger$ & $<\!10^{-4}$ \\
 & NQ    & SAPLMA (.847)          & $+.048$ $[+.030,+.065]^\dagger$ & $<\!10^{-4}$ \\
 & TQA   & SelfCheck-ngram (.895) & $+.033$ $[+.016,+.051]^\dagger$ & $<\!10^{-4}$ \\
\bottomrule
\end{tabular}}
\caption{Paired bootstrap of the AUROC difference between BEATRICE and the strongest of
all $23$ baselines on each column, sampling-based methods included ($10^4$ resamples of
the shared examples; two-sided $p$). The with-context pairing scores the original
contexts, the set every baseline scores. A dagger marks differences whose $95\%$
interval excludes zero.}
\label{tab:significance}
\end{table}

The sterner test is against our own strongest single signal rather than an external
baseline. Section~\ref{sec:anatomy} shows the hidden-state encoder, read on its own, is by
a wide margin the best single signal; Table~\ref{tab:champion-vs-hidden} asks whether fusing
the others onto it buys a significant gain. We bootstrap the paired AUROC difference between
the champion and the hidden-state encoder on the examples they share. The champion improves
on it in every column, and the gain is significant on both held-out single-hop sets in each
setting, largest there ($+0.016$ to $+0.021$ AUROC), while it is small and not significant
on the in-domain aggregate with context. The value of combining signals shows up where a
single signal is most brittle, out of distribution, which is exactly where a detector must
hold up.

\begin{table}[t]
\centering
\small
\setlength{\tabcolsep}{4pt}
\resizebox{\columnwidth}{!}{%
\begin{tabular}{ll cc c}
\toprule
Setting & Data & Champion & Hidden only & $\Delta$ (95\% CI) \\
\midrule
\multirow{4}{*}{Closed-book}
 & Test     & .839 & .832 & $+$.007$^\dagger$ [.000,\,.014] \\
 & PopQA    & .924 & .917 & $+$.007 [$-$.000,\,.014] \\
 & NQ       & .814 & .793 & $+$.021$^\dagger$ [.011,\,.032] \\
 & TQA      & .876 & .865 & $+$.010$^\dagger$ [.000,\,.020] \\
\midrule
\multirow{4}{*}{With context}
 & Test     & .926 & .921 & $+$.005 [$-$.001,\,.011] \\
 & PopQA    & .948 & .942 & $+$.006 [$-$.000,\,.012] \\
 & NQ       & .894 & .878 & $+$.016$^\dagger$ [.005,\,.027] \\
 & TQA      & .929 & .910 & $+$.019$^\dagger$ [.011,\,.028] \\
\bottomrule
\end{tabular}%
}
\caption{Paired significance of the fused champion over the strongest single signal, the
hidden-state encoder read on its own. For each evaluation set we bootstrap the AUROC
difference ($10{,}000$ resamples over the examples the two models share) and report the
$95\%$ interval; a dagger marks the columns whose interval excludes zero. The champion
improves on the hidden-state encoder in every column, and the improvement is significant on
both held-out single-hop sets (NQ, TriviaQA) in each setting and on the closed-book
in-domain test, while it is smallest and not significant on the in-domain aggregate with
context. Figures are computed on the shared examples (original passages), so the
with-context in-domain value is marginally below Table~\ref{tab:main-detection}.}
\label{tab:champion-vs-hidden}
\end{table}

\section{Answer-Span Localization}
\label{app:span}
Section~\ref{sec:anatomy} finds the read-out drawing mainly on the answer span.
Table~\ref{tab:span-ablation} gives the full breakdown: we zero one token region at a
time, context, question, or answer, retrain the detector, and measure the change in
AUROC. Removing the answer tokens costs the most by far in both settings, while removing
the question or the retrieved context costs little, so the detector reads correctness off
the model's handling of the answer rather than off the context tokens themselves.

\begin{table*}[t]
\centering
\small
\setlength{\tabcolsep}{4pt}
\begin{tabular}{l cccc cccc}
\toprule
& \multicolumn{4}{c}{Closed-book} & \multicolumn{4}{c}{With context} \\
\cmidrule(lr){2-5} \cmidrule(lr){6-9}
Dropped span & Test & PopQA & NQ & TQA & Test & PopQA & NQ & TQA \\
\midrule
None (full)   & .839/.913 & .924/.976 & .814/.848 & .874/.762 & .932/.826 & .948/.965 & .896/.941 & .930/.774 \\
Context       & --        & --        & --        & --        & .927/.809 & .944/.959 & .883/.929 & .925/.758 \\
Question      & .837/.914 & .920/.974 & .818/.848 & .882/.757 & .926/.800 & .943/.962 & .880/.931 & .927/.767 \\
Answer        & .831/.912 & .911/.971 & .803/.841 & .842/.714 & .919/.801 & .944/.963 & .884/.932 & .903/.711 \\
\bottomrule
\end{tabular}
\caption{Token-span ablation: encoders retrained with one span's features zeroed
at training time (evaluation protocol unchanged; each cell reports
\textbf{AUROC\,/\,AUPRC} with the incorrect answer as the AUPRC positive class).
Dropping the \emph{answer} span is consistently the most damaging cut in both
settings (with context: $-.013$ Test / $-.027$ TQA AUROC; closed-book: $-.013$
PopQA / $-.032$ TQA), while dropping the question span is nearly free --- and even
helps slightly on closed-book OOD --- indicating the correctness signal
concentrates on the answer tokens. Dropping the retrieved context costs little for
detection even in the with-context setting, consistent with the probe reading the
model's \emph{processing} of the context off the answer tokens rather than the
context tokens themselves.}
\label{tab:span-ablation}
\end{table*}

\section{How Far Surface Text Alone Goes}
\label{app:textonly}
BEATRICE reads a model's internal signals, but one might ask whether answer correctness
is already written on the surface, in the text of the question, the model's answer, and
the retrieved passage. If it were, a detector would not need the internal states at all.
We test this directly. On the same training split BEATRICE uses, and selected on the same
validation split, we train a ladder of text-only classifiers of increasing capacity:
linear models over TF-IDF features, an MLP over the same features, and a fine-tuned
transformer encoder, e5-base ($110$M parameters). Each reads only text: the question, the
model's final answer, and, with context, the retrieved passage.
Table~\ref{tab:textonly-control} reports their AUROC beside BEATRICE, whose two champions
have $3.1$M and $4.1$M parameters.

\begin{table*}[t]
\centering
\small
\setlength{\tabcolsep}{6pt}
\begin{tabular}{l cccc cccc}
\toprule
& \multicolumn{4}{c}{Closed-book} & \multicolumn{4}{c}{With context} \\
\cmidrule(lr){2-5} \cmidrule(lr){6-9}
& ID & \multicolumn{3}{c}{OOD} & ID & \multicolumn{3}{c}{OOD} \\
\cmidrule(lr){2-2} \cmidrule(lr){3-5} \cmidrule(lr){6-6} \cmidrule(lr){7-9}
Predictor & Test & PopQA & NQ & TQA & Test & PopQA & NQ & TQA \\
\midrule
\multicolumn{9}{l}{\emph{Text-only classifiers (question, answer, and retrieved passage as text)}} \\
Majority class            & .500 & .500 & .500 & .500 & .500 & .500 & .500 & .500 \\
TF-IDF + logistic regr.   & .785 & .774 & .676 & .643 & .901 & .863 & .819 & .807 \\
TF-IDF + linear SVM       & .746 & .760 & .658 & .629 & .869 & .836 & .787 & .765 \\
TF-IDF + MLP              & .790 & .781 & .664 & .652 & .902 & .857 & .833 & .807 \\
Fine-tuned e5-base (110M) & .831 & .849 & .771 & .765 & .900 & .814 & .620 & .637 \\
\midrule
\multicolumn{9}{l}{\emph{Internal signals (ours)}} \\
BEATRICE ($3.1$M\,/\,$4.1$M) & \textbf{.839} & \textbf{.924} & \textbf{.814} & \textbf{.874} & \textbf{.932} & \textbf{.948} & \textbf{.896} & \textbf{.930} \\
\bottomrule
\end{tabular}
\caption{Predicting answer correctness of Qwen3.5-9B from surface text alone, against
BEATRICE. Each cell is AUROC. Every text classifier is trained on the same training split
and selected on the same validation split as BEATRICE; its input is the question, the
model's final answer, and, in the with-context setting, the retrieved passage. In-domain
Test is the multi-hop aggregate; PopQA, NQ, and TQA are out-of-distribution. The two
figures after BEATRICE are its closed-book and with-context champion sizes; both are far
smaller than the fine-tuned encoder yet lead every column. Per column, \textbf{bold} is
best.}
\label{tab:textonly-control}
\end{table*}

Surface text carries a strong but incomplete signal. The text models come within a few
points of BEATRICE in-domain, reaching the low $.90$s with context, so correctness is far
from illegible on the surface. But BEATRICE leads every column, in-domain and out, and the
text models fall away on the harder out-of-distribution sets: with context on NQ and
TriviaQA the fine-tuned encoder drops into the low $.60$s, below the linear text models
despite its far greater capacity, a sign that it fits the in-domain distribution too
closely. Part of what a text model fits is a dataset-specific correlation between answer
form and correctness rather than correctness itself. Internal signals are not what makes
an in-domain detector work, but they are what keeps it reliable across distributions, and
they do so at a fraction of the size: BEATRICE's champions are $3.1$M and $4.1$M
parameters against the encoder's $110$M.

The softest column for BEATRICE is closed-book TriviaQA, and it is worth naming why. This
is the regime where the model already knows the answer: it is correct on $71\%$ of these
single-hop popular-entity questions, the highest rate among our datasets, so correctness
is legible from the answer form and from the model's own token confidence. It is also the
single column where sampling-based baselines overtake BEATRICE in the main comparison
(Table~\ref{tab:main-detection}), for the same reason. The effect is specific to the
closed-book regime: once a passage is present, BEATRICE leads TriviaQA by more than a
tenth of a point of AUROC.

\section{Interpretability Analyses}
\label{app:anatomy-signals}
This appendix looks inside the trained detector: which signal family determines
an encoder's decision, when the other families get to act, and what they respond
to. Every analysis queries the deployed encoders directly; nothing is retrained.

\paragraph{Which family decides.}
We ablate one family at a time inside a deployed combination encoder, replacing
its input channels with their means over the evaluation data, and record how
often the encoder's decision flips and how often the flip is in the right
direction (Table~\ref{tab:anatomy-flip-profile}). Hidden states dominate:
ablating them changes between a fifth and two thirds of all decisions, and with
context the decision they induce is almost always the correct one ($.964$ and
$.932$ across the two backbones). The auxiliary families each change only a few
percent of decisions, and which of them can be trusted switches with the
setting. Closed-book, token probabilities are the one precise auxiliary
($.746$ and $.714$) while attention flips are near coin flips; with context the
order reverses: residual-stream contributions ($.897$ and $.792$) and attention
distributions ($.809$ and $.733$) become precise, and token probabilities fall
back. The same pattern holds on both backbones.

\begin{figure}[t]
\centering
\includegraphics[width=\columnwidth]{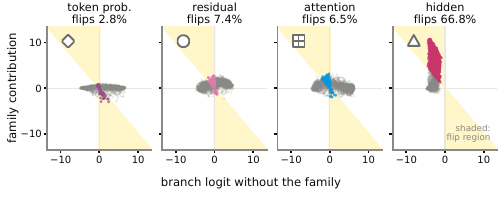}
\caption{When the auxiliary families change a decision, with-context branches
on Qwen3.5-9B, full test split. Each panel plots the encoder score with the
family ablated ($x$) against the family's contribution ($y$); in the shaded
wedge the contribution changes the decision, and colored points are the
flipped examples. Prob and Attn are measured in \lb\gH\gP\gA\rb{}, Resid in
\lb\gH\gP\gR\rb{}.}
\label{fig:boundary-band}
\end{figure}

\paragraph{When the auxiliaries act.}
Figure~\ref{fig:boundary-band} plots, for every with-context test example, the
encoder score with one family ablated against that family's contribution; in
the shaded wedge the contribution is large and opposed enough to change the
decision. The three auxiliary families form flat clouds: their contributions
stay within roughly one logit, so they reach the flip region only where the
rest of the encoder already sits near the boundary. Concretely, of the
decisions each auxiliary family flips, the share on which the remaining score
is within one logit of the boundary is $.79$ to $.94$ in the panels shown, and
between $.72$ and $1.00$ in eleven of the twelve
family--setting--backbone combinations overall; the one exception is Gemma's
closed-book residual family ($.45$), which is also the one auxiliary with poor
flip precision. Hidden states behave differently: their contribution is an
order of magnitude larger, and they overturn the other families' verdict at
any confidence. The auxiliary families therefore act as narrow-scope
corrections, re-deciding only the cases the rest of the encoder finds
uncertain, which is consistent with fusion's small but consistent gains in the
main results.

\begin{figure}[t]
\centering
\includegraphics[width=\columnwidth]{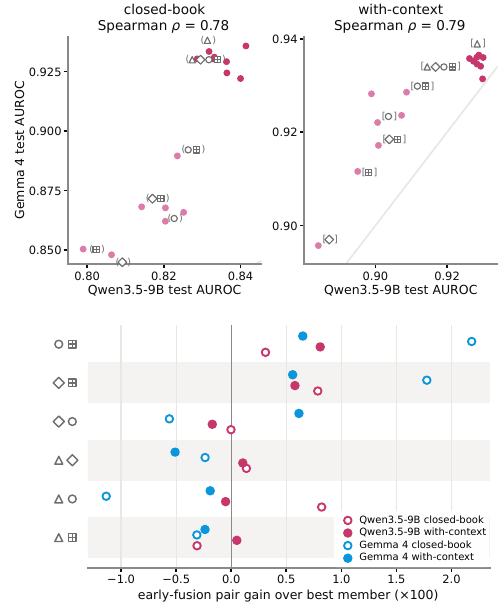}
\caption{Cross-backbone structure of the fifteen combination encoders, full
test split. Top: each encoder's AUROC on Qwen3.5-9B ($x$) against Gemma\,4
($y$), one panel per setting; dark points read hidden states, light ones do
not. Bottom: the AUROC of each two-family encoder minus the AUROC of the
better of its two single-family members ($\times 100$).}
\label{fig:cross-model-synergy}
\end{figure}

\paragraph{Which combinations pay off, on either backbone.}
Figure~\ref{fig:cross-model-synergy} compares all fifteen combination encoders
across the two backbones on the test split. Two regularities emerge. First, the
combination ranking is largely conserved: the Spearman rank correlation between
backbones is $.78$ closed-book and $.79$ with-context, and the encoders that
read hidden states form the top group on both. Second, early fusion is
super-additive only among the weak families. We score each two-family encoder
by its AUROC minus the AUROC of the better of its two single-family encoders.
The clearest case is token probabilities with attention: this gain is positive
in seventeen of the twenty combinations of two settings, two backbones, and
five evaluation splits (median $0.8$ AUROC points, up to $6.4$ on closed-book
TriviaQA); residual with attention shows a weaker version of the same
tendency. Pairs that contain
hidden states hover at zero gain, and pairing the two numeric families rarely
helps. The interactions worth having thus live among the weak families, while
hidden states neither need nor gain from sharing an encoder; both facts, and
the ranking itself, are properties of the signal families rather than of one
backbone.

\paragraph{What the attention family responds to.}
The attention family endorses answers that have verbatim support in the
retrieved passage, so its reliability is set by how well support tracks
correctness. Table~\ref{tab:anatomy-grounding} traces this across three
evaluation sets. On the in-domain test split retrieval is good: the passage
lifts the generator from $29.4\%$ to $78.4\%$ accuracy, support and
correctness mostly agree, and endorsing support is usually right (flip
precision $.809$). On NQ retrieval is often harmful: accuracy falls when the
passage is added ($42.7 \rightarrow 32.5$), $45.1\%$ of wrong answers are
copied verbatim from the passage, and the family's precision drops to $.672$.
The subset rows locate the failures. Wherever an answer appears in the passage
but is wrong, the family's flips are wrong essentially always ($.111$
in-domain, $.000$ elsewhere): it endorses support, not truth. The reverse
mismatch, a correct answer without support, becomes systematic only on
TriviaQA ($.040$), the one set the generator mostly answers from memory
($70.8\%$ closed-book); on the other two sets the family reacts weakly to
missing support, and its rare flips there are unsystematic ($.613$, $.500$).
Both failure modes are one mechanism seen under different retrieval
distributions: the family reads verbatim support, and support is only as good
a proxy for correctness as the retriever makes it.

\begin{figure}[t]
\centering
\includegraphics[width=\columnwidth]{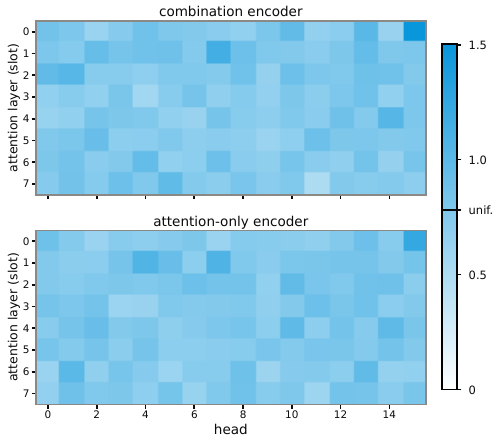}
\caption{Attribution share of each layer--head cell of the attention family
on Qwen3.5-9B: absolute integrated-gradients attribution of the encoder score,
summed over that head's channels and all spans, $n{=}400$ test examples. Top:
the with-context combination branch \lb\gH\gP\gA\rb{}; bottom: the
single-family encoder \lb\gA\rb{}. The mark on the colorbar is the uniform
share.}
\label{fig:attention-heads}
\end{figure}

\paragraph{No privileged attention heads.}
The attention family's channels are per-head summaries, so one can ask whether
a few heads carry the signal. To measure each head's influence we attribute
the encoder's score to its input channels with integrated gradients, moving
from a baseline in which the family's channels are set to their means over the
evaluation data up to the actual input; a head's value is the absolute
attribution of its channels, summed over tokens, spans and $400$ test
examples, and normalised so that all heads sum to one.
Figure~\ref{fig:attention-heads} shows the resulting share of every
layer--head cell, in the with-context combination branch and in the encoder
that reads attention alone. Both maps are close to
uniform: the eight largest cells hold $8.6\%$ and $8.3\%$ of the attribution
against a uniform $6.3\%$, and in the combination branch ablating those eight
heads' channels perturbs the encoder no more than ablating eight random ones.
What the probe reads from attention is spread redundantly across heads rather
than localised in a few.

\begin{table}[t]
\centering
\footnotesize
\setlength{\tabcolsep}{5pt}
\begin{tabular}{l cc cc}
\toprule
& \multicolumn{2}{c}{Closed-book} & \multicolumn{2}{c}{With-context} \\
\cmidrule(lr){2-3}\cmidrule(lr){4-5}
& flip rate & flip precision & flip rate & flip precision \\
\midrule
\multicolumn{5}{l}{\emph{Qwen3.5-9B}} \\
\gH{} Hidden & .385 & .598 {\scriptsize[.560,\,.641]}
             & .668 & .964 {\scriptsize[.956,\,.972]} \\
\gP{} Prob   & .039 & \textbf{.746} {\scriptsize[.644,\,.864]}
             & .028 & .583 {\scriptsize[.488,\,.690]} \\
\gR{} Resid  & .120 & .556 {\scriptsize[.489,\,.622]}
             & .074 & \textbf{.897} {\scriptsize[.857,\,.933]} \\
\gA{} Attn   & .034 & .569 {\scriptsize[.431,\,.706]}
             & .065 & \textbf{.809} {\scriptsize[.758,\,.861]} \\
\midrule
\multicolumn{5}{l}{\emph{Gemma\,4}} \\
\gH{} Hidden & .173 & .413 {\scriptsize[.351,\,.471]}
             & .626 & .932 {\scriptsize[.915,\,.947]} \\
\gP{} Prob   & .014 & \textbf{.714} {\scriptsize[.524,\,.905]}
             & .020 & .600 {\scriptsize[.433,\,.767]} \\
\gR{} Resid  & .073 & .227 {\scriptsize[.145,\,.300]}
             & .083 & \textbf{.792} {\scriptsize[.720,\,.864]} \\
\gA{} Attn   & .019 & .310 {\scriptsize[.138,\,.483]}
             & .050 & \textbf{.733} {\scriptsize[.627,\,.827]} \\
\bottomrule
\end{tabular}
\caption{Decision flips under family ablation, full test split. One signal
family's input channels are replaced by their means over the evaluation data;
a flip means the encoder's decision, $P(\text{answer correct})$ above or below
one half, changes. Flip rate is the fraction of examples flipped; flip
precision is the fraction of flips on which the intact encoder decides
correctly, with $95\%$ bootstrap intervals. Closed-book rows are measured in
\lp\gH\gP\gA\rp{}, with-context rows in \lb\gH\gP\gA\rb{}; the Resid rows use
\lp\gH\gR\rp{} and \lb\gH\gP\gR\rb{}. \textbf{Bold} marks the most precise
auxiliary family per setting. Gemma closed-book precisions should be read
against that split's $9\%$ positive rate.}
\label{tab:anatomy-flip-profile}
\end{table}

\begin{table}[t]
\centering
\footnotesize
\setlength{\tabcolsep}{5pt}
\resizebox{\columnwidth}{!}{%
\begin{tabular}{>{\raggedright\arraybackslash}p{0.58\columnwidth} ccc}
\toprule
& test & NQ & TriviaQA \\
\midrule
Generator answer accuracy (\%) & & & \\
\quad closed-book & $29.4$ & $42.7$ & $70.8$ \\
\quad with the retrieved passage & $78.4$ & $32.5$ & $78.1$ \\
\addlinespace[2pt]
Wrong answers present in passage & $34.0\%$ & $45.1\%$ & $34.7\%$ \\
\midrule
Attention flip precision & $.809$ & $.672$ & $.607$ \\
\quad answer in passage, but wrong & $.111$ & $.000$ & $.000$ \\
\quad answer correct, not in passage & $.613$ & $.500$ & $.040$ \\
\bottomrule
\end{tabular}}
\caption{The attention family against retrieval quality, on the with-context
branch \lb\gH\gP\gA\rb{} of Qwen3.5-9B. The first three rows characterise the
data: the generator's answer accuracy without and with the retrieved passage,
and the share of wrong answers whose text nevertheless appears verbatim in the
passage (normalised substring match). The last three rows are the
decision-flip precision of Table~\ref{tab:anatomy-flip-profile}, computed on
all examples and then restricted to the two subsets on which verbatim support
and correctness disagree.}
\label{tab:anatomy-grounding}
\end{table}

\section{Detection Under Real Retrieved Context}
\label{app:realrag}
The with-context results in the main text score detectors on the datasets' own supporting
passages. Before turning the detector into a controller, we check that it still detects
correctness when the context comes from real retrieval, the condition a deployed system
faces. We build a $3{,}000$-question test set, $500$ from each of the six datasets (the
first $500$ of each canonical test split, disjoint from probe training), and retrieve
$k{=}3$ passages per question with E5 over Wikipedia-2018, the same retriever the controller
of Section~\ref{sec:pgir} uses. The backbone answers from these passages, and each answer is
labeled by the majority vote of the three judge prompts (Appendix~\ref{app:data}) without
the consensus-purification step, so all $3{,}000$ questions are kept, including the harder
cases where the judges disagree. Every trained detector, ours included, reuses its existing
with-context weights and is evaluated zero-shot on this set; we report per-dataset AUROC and
AUPRC on both backbones (Qwen3.5-9B, with a retrieval-augmented accuracy of $48\%$, and
Gemma\,4, $37\%$).

Tables~\ref{tab:realrag-detection} and~\ref{tab:realrag-detection-gemma} show the ranking is
stable. BEATRICE is first on both backbones, at $0.891$ on Qwen ($+0.039$ over the strongest
baseline) and $0.911$ on Gemma ($+0.064$), and it leads most individual datasets, four of six
on Qwen and all six on Gemma. The families behave as in the gold-context comparison:
single-pass probes are the strongest baselines, and on Gemma the sampling scores again fall
below chance, the consistent-refusal effect of Appendix~\ref{sec:appendix-gemma}. Reading
correctness from internal signals thus transfers from gold to retrieved context without
retraining.

\begin{table*}[t]
\centering
\footnotesize
\setlength{\tabcolsep}{4pt}
\begin{tabular}{l ccc ccc c}
\toprule
& \multicolumn{3}{c}{Single-hop} & \multicolumn{3}{c}{Multi-hop} & \\
\cmidrule(lr){2-4} \cmidrule(lr){5-7}
Method & PopQA & NQ & TQA & HotpotQA & 2Wiki & MuSiQue & All \\
\midrule
\emph{Random AUPRC (error rate)} & \emph{.504} & \emph{.396} & \emph{.158} & \emph{.492} & \emph{.706} & \emph{.882} & \emph{.523} \\
\midrule
BEATRICE & \textbf{.866}/\textbf{.868} & \textbf{.836}/\textbf{.781} & .899/\textbf{.696} & \textbf{.864}/\textbf{.866} & \textbf{.832}/\textbf{.909} & .798/.964 & \textbf{.891}/\textbf{.894} \\
SAPLMA (mean) & .813/.809 & .770/.644 & \textbf{.910}/.643 & .826/.780 & .757/.865 & \textbf{.811}/\textbf{.969} & .852/.836 \\
SAPLMA (last) & .811/.799 & .778/.653 & .844/.566 & .790/.774 & .775/.892 & .766/.958 & .836/.827 \\
LLMsKnow & .800/.789 & .756/.656 & .815/.562 & .795/.798 & .718/.843 & .713/.946 & .812/.813 \\
SelfCheckGPT-ngram & .761/.759 & .779/.719 & .862/.596 & .826/.808 & .689/.829 & .723/.949 & .805/.812 \\
Lookback-Lens & .725/.753 & .699/.573 & .784/.460 & .757/.746 & .703/.841 & .679/.933 & .748/.757 \\
SEP & .680/.679 & .585/.458 & .726/.367 & .688/.664 & .646/.779 & .604/.914 & .702/.687 \\
Bayesian SE & .655/.611 & .668/.520 & .791/.330 & .723/.659 & .557/.724 & .627/.923 & .683/.647 \\
Semantic Entropy (discrete) & .653/.610 & .666/.520 & .790/.325 & .720/.655 & .557/.724 & .623/.922 & .681/.645 \\
Semantic Entropy & .650/.615 & .660/.493 & .787/.319 & .710/.625 & .561/.737 & .622/.922 & .677/.630 \\
SelfCheckGPT-NLI & .622/.680 & .708/.646 & .787/.517 & .742/.739 & .525/.736 & .546/.908 & .673/.711 \\
EigenScore & .567/.536 & .686/.584 & .701/.298 & .706/.679 & .594/.741 & .620/.918 & .663/.657 \\
LLM-Check (combined) & .613/.642 & .642/.575 & .674/.472 & .675/.704 & .598/.782 & .630/.918 & .647/.702 \\
ReDeEP & .609/.586 & .617/.494 & .619/.241 & .638/.603 & .535/.702 & .600/.897 & .646/.628 \\
ReDeEP-PKS & .613/.585 & .615/.487 & .616/.235 & .639/.608 & .537/.703 & .597/.896 & .643/.626 \\
LLM-Check (entropy) & .539/.588 & .614/.557 & .590/.322 & .625/.660 & .513/.739 & .617/.923 & .599/.660 \\
ReDeEP-ECS & .561/.569 & .582/.476 & .603/.318 & .547/.540 & .531/.719 & .588/.915 & .599/.602 \\
LUMINA-IPR & .535/.544 & .575/.449 & .555/.222 & .641/.664 & .521/.725 & .538/.903 & .582/.610 \\
LLM-Check (hidden) & .529/.550 & .547/.461 & .554/.284 & .560/.593 & .503/.743 & .648/.930 & .579/.623 \\
LUMINA & .543/.543 & .574/.448 & .551/.223 & .611/.647 & .501/.720 & .537/.904 & .569/.607 \\
LLM-Check (perplexity) & .585/.615 & .549/.449 & .579/.273 & .496/.525 & .587/.798 & .464/.876 & .527/.586 \\
LLM-Check (attention) & .515/.531 & .537/.434 & .531/.178 & .524/.495 & .564/.738 & .495/.889 & .519/.540 \\
LUMINA-MMD & .520/.530 & .546/.441 & .499/.149 & .444/.431 & .503/.726 & .544/.899 & .504/.511 \\
\bottomrule
\end{tabular}
\caption{Real-retrieval detection on \textbf{Qwen3.5-9B}: AUROC\,/\,AUPRC on each dataset under real retrieved context (E5 over Wikipedia-2018, $k{=}3$), $500$ questions each. AUPRC positive class is the incorrect answer; the italic row is the random-guess AUPRC (per-column error rate). \textbf{Bold} marks the best in each column per metric. Every trained detector reuses its with-context weights and is evaluated zero-shot.}
\label{tab:realrag-detection}
\end{table*}

\begin{table*}[t]
\centering
\footnotesize
\setlength{\tabcolsep}{4pt}
\begin{tabular}{l ccc ccc c}
\toprule
& \multicolumn{3}{c}{Single-hop} & \multicolumn{3}{c}{Multi-hop} & \\
\cmidrule(lr){2-4} \cmidrule(lr){5-7}
Method & PopQA & NQ & TQA & HotpotQA & 2Wiki & MuSiQue & All \\
\midrule
\emph{Random AUPRC (error rate)} & \emph{.562} & \emph{.440} & \emph{.266} & \emph{.680} & \emph{.882} & \emph{.930} & \emph{.627} \\
\midrule
BEATRICE & \textbf{.877}/\textbf{.912} & \textbf{.819}/\textbf{.802} & \textbf{.883}/\textbf{.805} & \textbf{.922}/\textbf{.963} & \textbf{.943}/\textbf{.992} & \textbf{.892}/\textbf{.991} & \textbf{.911}/\textbf{.950} \\
LLMsKnow & .765/.845 & .696/.690 & .749/.672 & .850/.925 & .847/.977 & .841/.985 & .847/.916 \\
SAPLMA (mean) & .782/.823 & .709/.673 & .739/.488 & .824/.903 & .800/.963 & .727/.970 & .806/.864 \\
SAPLMA (last) & .781/.842 & .665/.620 & .699/.478 & .830/.911 & .791/.965 & .799/.981 & .799/.872 \\
LUMINA & .711/.795 & .648/.623 & .615/.484 & .768/.889 & .815/.971 & .795/.981 & .761/.863 \\
LUMINA-IPR & .756/.837 & .649/.600 & .712/.579 & .765/.890 & .753/.952 & .756/.972 & .761/.855 \\
LUMINA-MMD & .707/.788 & .644/.602 & .612/.442 & .753/.873 & .806/.969 & .789/.980 & .755/.847 \\
Lookback-Lens & .749/.820 & .662/.602 & .680/.462 & .763/.878 & .739/.943 & .789/.978 & .745/.827 \\
LLM-Check (perplexity) & .753/.746 & .669/.558 & .705/.462 & .771/.859 & .730/.941 & .696/.953 & .744/.791 \\
LLM-Check (hidden) & .714/.808 & .661/.625 & .677/.584 & .704/.851 & .701/.948 & .635/.951 & .718/.829 \\
LLM-Check (entropy) & .766/.788 & .674/.589 & .632/.391 & .716/.819 & .674/.909 & .704/.960 & .710/.764 \\
SelfCheckGPT-ngram & .726/.741 & .613/.566 & .801/.539 & .716/.819 & .609/.908 & .475/.928 & .696/.768 \\
LLM-Check (combined) & .734/.807 & .671/.648 & .515/.337 & .609/.801 & .653/.932 & .572/.952 & .640/.776 \\
ReDeEP & .612/.638 & .545/.455 & .556/.298 & .655/.771 & .629/.905 & .630/.960 & .630/.702 \\
ReDeEP-PKS & .610/.637 & .546/.456 & .556/.295 & .654/.769 & .617/.899 & .621/.960 & .627/.698 \\
LLM-Check (attention) & .525/.556 & .545/.469 & .524/.269 & .589/.702 & .598/.889 & .636/.945 & .563/.646 \\
SEP & .528/.603 & .553/.479 & .569/.335 & .536/.722 & .624/.911 & .522/.934 & .550/.668 \\
ReDeEP-ECS & .540/.566 & .497/.436 & .519/.274 & .533/.677 & .558/.875 & .599/.950 & .549/.641 \\
EigenScore & .589/.605 & .611/.493 & .551/.282 & .534/.684 & .545/.891 & .437/.915 & .531/.618 \\
Bayesian SE & .584/.626 & .490/.428 & .602/.349 & .451/.676 & .394/.860 & .367/.916 & .470/.618 \\
Semantic Entropy (discrete) & .585/.625 & .489/.427 & .602/.349 & .451/.675 & .395/.860 & .366/.916 & .470/.617 \\
Semantic Entropy & .583/.627 & .485/.420 & .600/.350 & .451/.671 & .394/.861 & .368/.915 & .469/.614 \\
SelfCheckGPT-NLI & .495/.670 & .491/.533 & .458/.371 & .238/.652 & .110/.851 & .218/.901 & .306/.631 \\
\bottomrule
\end{tabular}
\caption{Real-retrieval detection on \textbf{Gemma\,4}, conventions as in Table~\ref{tab:realrag-detection}. On this backbone the sampling detectors fall to or below chance (the consistent-refusal effect of Appendix~\ref{sec:appendix-gemma}).}
\label{tab:realrag-detection-gemma}
\end{table*}

\section{PGIR End-Task Results}
\label{app:pgir-results}
Table~\ref{tab:pgir-application} gives the full end-task comparison summarized in
Section~\ref{sec:pgir}: exact-match and F1 on each dataset, macro means, and generations
per query, for PGIR and the confidence-control baselines under one shared backbone and
retriever.

\begin{table*}[t]
\centering
\small
\setlength{\tabcolsep}{4pt}
\begin{tabular}{l ccc ccc cc c}
\toprule
& \multicolumn{3}{c}{In-domain (multi-hop)} & \multicolumn{3}{c}{Out-of-distribution} & \multicolumn{2}{c}{Macro} & \\
\cmidrule(lr){2-4} \cmidrule(lr){5-7} \cmidrule(lr){8-9}
Method & HotpotQA & 2WikiMQA & MuSiQue & PopQA & NQ & TriviaQA & EM & F1 & Gen \\
\midrule
No-RAG          & .220/.293 & .238/.282 & .034/.114 & .192/.246 & .176/.267 & .574/.617 & .239 & .303 & 1.00 \\
Vanilla RAG     & .322/.437 & .260/.309 & .060/.133 & \underline{.447}/\underline{.526} & \underline{.371}/\underline{.485} & \underline{.757}/\textbf{.818} & .370 & .451 & 1.00 \\
\midrule
SKR             & .310/.416 & .296/.342 & .052/.126 & .440/.519 & .364/.474 & .738/.796 & .367 & .445 & 1.00 \\
Adaptive-RAG    & .322/.431 & .282/.323 & .060/.098 & .433/.496 & .319/.425 & .718/.769 & .356 & .424 & 1.63 \\
DRAGIN          & .300/.394 & .298/.343 & .062/.151 & .415/.473 & .354/.467 & .672/.724 & .350 & .425 & 3.55 \\
SeaKR           & \textbf{.390}/\textbf{.509} & \underline{.340}/\textbf{.434} & \textbf{.092}/\underline{.173} & .296/.353 & .261/.372 & .695/.765 & .346 & .434 & 13.63 \\
Probing-RAG     & \underline{.364}/\underline{.477} & .322/.384 & .078/\textbf{.179} & .422/.492 & .346/.462 & .731/.780 & \underline{.377} & \underline{.462} & 2.56 \\
\midrule
PGIR (ours)     & .332/.435 & \textbf{.356}/\underline{.403} & \underline{.090}/.172 & \textbf{.489}/\textbf{.554} & \textbf{.420}/\textbf{.527} & \textbf{.758}/\underline{.812} & \textbf{.408} & \textbf{.484} & 3.56 \\
\bottomrule
\end{tabular}
\caption{PGIR as a downstream controller, against the confidence-control family:
systems that gate or verify retrieval with a self-assessed signal, all sharing one
backbone and retriever. Cells are exact-match\,/\,F1. Macro is the six-set mean and Gen is
generations per query. Per column and metric, \textbf{bold}~$=$~best and
\underline{underline}~$=$~second.}
\label{tab:pgir-application}
\end{table*}

\section{Case Studies}
\label{app:cases}
Two forward passes, two decisions. Table~\ref{tab:case-studies} traces questions from the
PGIR runs on both backbones and shows the same probe auditing and deciding. When the closed-book draft is
confident and correct, the pre-retrieval gate accepts it and skips retrieval. When it is a
confident but wrong guess, the probe scores it low and PGIR retrieves; a grounded draft
that clears the threshold is then trusted, even when it overturns the closed-book answer.
Asked what gemstone ``The Moonstone'' is in Wilkie Collins's novel, the model first echoes
the title, ``Moonstone'', which the probe scores low ($0.57$); with the passage it answers
``Diamond'' and the score rises to $0.90$, so PGIR returns it. When retrieval does not
help, no draft clears the threshold and PGIR returns its best candidate without trusting
it, leaving the wrong answer flagged.

\newcommand{\cbwc}[2]{#1 \small(#2)}
\begin{table*}[t]
\centering
\small
\setlength{\tabcolsep}{5pt}
\begin{tabular}{>{\raggedright\arraybackslash}p{4.1cm} >{\raggedright\arraybackslash}p{2.9cm} >{\raggedright\arraybackslash}p{2.9cm} >{\raggedright\arraybackslash}p{1.9cm} >{\raggedright\arraybackslash}p{2.4cm}}
\toprule
Question & Closed-book answer (score) & With-context answer (score) & Gold & PGIR action \\
\midrule
\multicolumn{5}{l}{\emph{Qwen3.5-9B}} \\
The Nun is based on what film series?
 & \cbwc{The Conjuring}{.67} & \cbwc{\emph{not retrieved}}{} & The Conjuring
 & Accept, no retrieval \\
\addlinespace
What gemstone is \emph{The Moonstone} in the novel by Wilkie Collins?
 & \cbwc{Moonstone}{.57} & \cbwc{Diamond}{.90} & Diamond
 & Retrieve, then trust \\
\addlinespace
Arctic King, Saladin, and Tom Thumb are which vegetable?
 & \cbwc{Onions}{.26} & \cbwc{Cauliflower}{.01} & Lettuce
 & Retrieve; both flagged, none trusted \\
\midrule
\multicolumn{5}{l}{\emph{Gemma 4}} \\
Who played Clayton Farlowe in \emph{Dallas}?
 & \cbwc{Michael Landon}{.05} & \cbwc{Howard Keel}{.65} & Howard Keel
 & Retrieve, then trust \\
\addlinespace
Which country is Haya bint Hussein's husband from?
 & \cbwc{Jordan}{.05} & \cbwc{United Arab Emirates}{.65} & United Arab Emirates
 & Retrieve, then trust \\
\addlinespace
What genre is the band \emph{Darkness}?
 & \cbwc{Horror}{.35} & \cbwc{Dark fantasy}{.09} & hard rock
 & Retrieve; both flagged, none trusted \\
\bottomrule
\end{tabular}
\caption{Case studies from the PGIR runs on both backbones (E5 over Wikipedia). Each score
is the calibrated $P(\text{answer correct})$ the controller acts on; the acceptance
threshold is $\tau{=}0.6$. A high closed-book score is accepted without retrieval; a low
one triggers retrieval, after which a draft clearing $\tau$ is trusted even when it
overturns the closed-book guess, while a pool that stays low is left flagged and the best
candidate is returned without claiming it is right.}
\label{tab:case-studies}
\end{table*}

\section{Generality Across Base Models}
\label{sec:appendix-gemma}
\begin{figure}[t]
\centering
\includegraphics[width=\columnwidth]{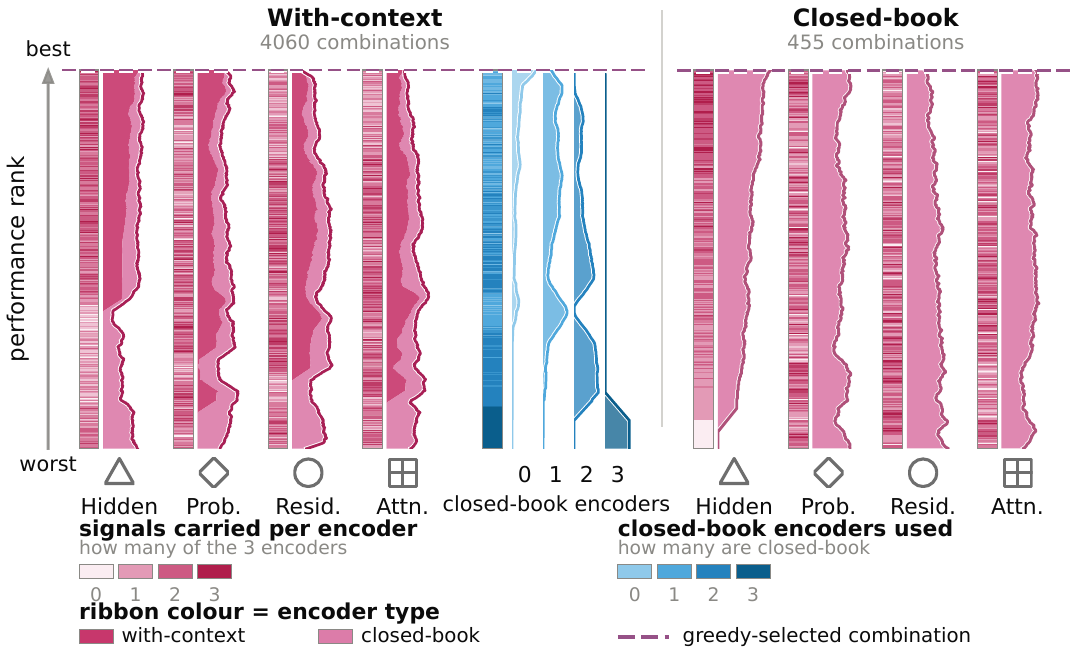}
\caption{Exhaustive ranking of all three-encoder combinations on Gemma\,4, best to worst,
in the format of Figure~\ref{fig:combination-search}. All-closed-book combinations sit at
the bottom in the with-context pool; the greedy-selected combination is marked by the
dashed line.}
\label{fig:gemma-combination-search}
\end{figure}

To test whether the detector and its selection protocol transfer beyond a single
base model, we repeat the entire pipeline on Gemma\,4 E4B (instruction-tuned), a
model that differs from Qwen3.5-9B in scale, architecture family, and attention
layout (only 7 of its 42 layers use full attention; the attention distribution
features read exactly those layers). Every protocol decision is held fixed: the
same data scale and split construction, correctness labels rebuilt from the new
model's own greedy answers under the same consensus purification, the same four
signal families with the read-out layer re-selected on validation, the same
encoder pool, and the same combination search with every choice made on the
validation split only. Greedy selection again reaches the global optimum of the
exhaustive enumeration over all three-encoder subsets in both settings (455
closed-book and 4{,}060 with-context candidates). Figure~\ref{fig:gemma-combination-search}
draws this landscape as Figure~\ref{fig:combination-search} does for Qwen. The two ends
agree: on both models the worst with-context combinations are built entirely from
closed-book encoders. The top does not. Reading the closed-book-count panel up toward the
optimum, Qwen concentrates on exactly one closed-book encoder, its ten best combinations all
carry one, whereas Gemma moves the other way, the zero-closed-book fraction rises sharply
and the one-closed-book fraction falls off, until nine of Gemma's ten best combinations
carry none. The closed-book prior that gives Qwen a small lift under retrieval does not
carry over: at Gemma's optimum the with-context signal is self-sufficient, and the value of
that prior is genuinely base-model dependent.

Table~\ref{tab:gemma-detection} reports the comparison in the format of
Table~\ref{tab:main-detection}. Under the validation selection protocol BEATRICE
leads both settings (closed-book validation AUROC .952 against .944 for the
strongest baseline; with-context .941 against .929), and it is best in five of
the eight evaluation columns, including all PopQA and NQ columns. The remaining
three columns go to a baseline by at most .014 AUROC: SAPLMA on closed-book NQ,
Lookback-Lens on the with-context in-domain test, and ReDeEP on with-context
TriviaQA. Baselines are markedly stronger on Gemma than on Qwen, so we report
this as a protocol-level win with per-column trade-offs rather than uniform
dominance.

Two systematic shifts are themselves findings. First, sampling-based consistency
methods collapse to or below chance in most columns: repeated samples of this
instruction-tuned model frequently agree on the same refusal or the same wrong
short answer (two thirds of queries collapse to a single semantic cluster, and
those queries have a higher error rate), so agreement across samples no longer
tracks correctness. Second, attention-carried signals are systematically
stronger on Gemma: Lookback-Lens rises from the .78 to .91 AUROC range on Qwen
to .92 to .97 here, and our own attention encoder shows the same shift. Internal-state
detectors remain strong across both base models while behavior-based scores do
not, which supports reading the model rather than sampling it.

Finally, Table~\ref{tab:gemma-pgir} repeats the application experiment of
Table~\ref{tab:pgir-application} on the Gemma backbone: same retriever and
evaluation splits, all trained controllers retrained for the new model, and
PGIR driven by the Gemma probe with its threshold and budget re-tuned on the
dev split only. The ranking of the confidence-control family is stable across
backbones: PGIR is first on macro EM and F1 and Probing-RAG second on both,
while individual sets again go to specialists (SeaKR on HotpotQA, at 17.5
generations per query against PGIR's 4.6, and Probing-RAG on TriviaQA).
\begin{table*}[t]
\centering
\small
\setlength{\tabcolsep}{4pt}
\begin{tabular}{l cccc cccc}
\toprule
& \multicolumn{4}{c}{Closed-book} & \multicolumn{4}{c}{With context} \\
\cmidrule(lr){2-5} \cmidrule(lr){6-9}
& ID & \multicolumn{3}{c}{OOD} & ID & \multicolumn{3}{c}{OOD} \\
\cmidrule(lr){2-2} \cmidrule(lr){3-5} \cmidrule(lr){6-6} \cmidrule(lr){7-9}
Method & Test & PopQA & NQ & TQA & Test & PopQA & NQ & TQA \\
\midrule
\multicolumn{9}{l}{\emph{Sampling-based (10 extra generations per query)}} \\
Semantic Entropy    & .339/.894 & .523/.888 & .525/.746 & .756/.717 & .595/.404 & .456/.648 & .384/.726 & .466/.428 \\
Bayesian SE         & .323/.891 & .524/.888 & .513/.737 & .749/.713 & .591/.403 & .449/.647 & .367/.720 & .451/.425 \\
SelfCheckGPT-NLI    & .199/.880 & .330/.883 & .485/.777 & .693/.729 & .390/.419 & .087/.628 & .169/.740 & .185/.448 \\
SelfCheckGPT-ngram  & .509/.921 & .832/.951 & .633/.790 & .832/.752 & .741/.568 & .821/.877 & .596/.817 & .779/.691 \\
EigenScore          & .458/.893 & .497/.862 & .603/.761 & .510/.450 & .742/.504 & .762/.784 & .632/.817 & .666/.546 \\
\midrule
\multicolumn{9}{l}{\emph{Single-signal, single forward pass}} \\
SAPLMA              & \underline{.929}/\underline{.992} & .937/\underline{.990} & \textbf{.866}/\textbf{.936} & \underline{.865}/\underline{.842} & .908/.841 & \underline{.963}/\underline{.984} & .908/.969 & .939/.939 \\
LLMsKnow            & .909/.990 & .930/.988 & .827/.917 & .822/.784 & .918/.867 & .955/.978 & .904/.965 & .942/.940 \\
SEP                 & .433/.897 & .459/.832 & .446/.644 & .534/.450 & .649/.429 & .747/.828 & .689/.848 & .637/.593 \\
LLM-Check           & .880/.987 & \underline{.940}/.989 & .820/.905 & .853/.825 & .876/.804 & .956/.981 & .884/.958 & .898/.897 \\
\midrule
\multicolumn{9}{l}{\emph{RAG-specific detectors (single forward pass)}} \\
Lookback-Lens       & -- & -- & -- & -- & \textbf{.948}/\textbf{.946} & .954/.982 & \underline{.939}/\textbf{.983} & \underline{.971}/\textbf{.980} \\
ReDeEP-PKS          & -- & -- & -- & -- & .922/\underline{.924} & .955/.982 & .934/\underline{.982} & \textbf{.974}/\underline{.979} \\
LUMINA              & -- & -- & -- & -- & .879/.819 & .934/.951 & .904/.966 & .935/.875 \\
\midrule
\multicolumn{9}{l}{\emph{Multi-signal fusion (ours)}} \\
BEATRICE            & \textbf{.936}/\textbf{.993} & \textbf{.957}/\textbf{.993} & \underline{.853}/\underline{.929} & \textbf{.879}/\textbf{.859} & \underline{.935}/.893 & \textbf{.972}/\textbf{.988} & \textbf{.946}/\textbf{.983} & .963/.961 \\
\bottomrule
\end{tabular}
\caption{Detecting incorrect answers of Gemma\,4 E4B (instruction-tuned) in the
closed-book and retrieval-augmented settings, mirroring
Table~\ref{tab:main-detection}: each cell reports AUROC\,/\,AUPRC with ``the answer is
wrong'' as the AUPRC positive class. The entire pipeline is re-run on the new base
model under the protocol of Table~\ref{tab:main-detection}: same data scale and
splits, consensus-purified labels regenerated from this model's own greedy answers,
all layers and hyperparameters of every method re-selected on the validation split
only, and the with-context evaluation again merging original and rewritten contexts.
The selected probes are
\lp{}\gH{}\gP{}\gA{}\rp{}\lp{}\gH{}\rp{}\lp{}\gH{}\gP{}\gR{}\gA{}\rp{} (closed-book) and
\lb{}\gA{}\rb{}\lb{}\gH{}\gP{}\rb{}\lb{}\gH{}\rb{} (with context). Sampling-based
scores fall to or below chance on this base model; see the discussion in the text.
Per column and metric, \textbf{bold}~$=$~best and \underline{underline}~$=$~second;
``--'' marks detectors that require retrieved context and are undefined closed-book.}
\label{tab:gemma-detection}
\end{table*}

\begin{table*}[t]
\centering
\small
\setlength{\tabcolsep}{4pt}
\begin{tabular}{l ccc ccc cc c}
\toprule
& \multicolumn{3}{c}{In-domain (multi-hop)} & \multicolumn{3}{c}{Out-of-distribution} & \multicolumn{2}{c}{Macro} & \\
\cmidrule(lr){2-4} \cmidrule(lr){5-7} \cmidrule(lr){8-9}
Method & HotpotQA & 2WikiMQA & MuSiQue & PopQA & NQ & TriviaQA & EM & F1 & Gen \\
\midrule
No-RAG          & .050/.069 & .006/.012 & .000/.003 & .048/.052 & .030/.052 & .298/.340 & .072 & .088 & 1.00 \\
Vanilla RAG     & .190/.298 & .086/.177 & .030/.078 & .375/\underline{.463} & .284/.396 & .602/.692 & .261 & .351 & 1.00 \\
\midrule
SKR             & .190/.297 & .112/.196 & .028/.074 & .368/.456 & .280/.392 & .602/.690 & .263 & .351 & 1.00 \\
Adaptive-RAG    & .222/.329 & .104/.194 & \textbf{.040}/\textbf{.097} & .364/.431 & \underline{.307}/\underline{.415} & .625/.700 & .277 & .361 & 1.70 \\
DRAGIN          & .164/.237 & .154/.201 & .020/.062 & .228/.284 & .225/.323 & .514/.584 & .218 & .282 & 2.64 \\
SeaKR           & \textbf{.264}/\textbf{.372} & \underline{.216}/\underline{.264} & .024/.070 & .223/.274 & .157/.239 & .544/.621 & .238 & .307 & 17.48 \\
Probing-RAG     & .226/.334 & .176/.230 & \textbf{.040}/.091 & \underline{.378}/.462 & .293/.410 & \textbf{.708}/\textbf{.779} & \underline{.304} & \underline{.384} & 3.43 \\
\midrule
PGIR (ours)     & \underline{.244}/\underline{.344} & \textbf{.260}/\textbf{.309} & \underline{.036}/\underline{.096} & \textbf{.397}/\textbf{.473} & \textbf{.326}/\textbf{.433} & \underline{.651}/\underline{.724} & \textbf{.319} & \textbf{.397} & 4.63 \\
\bottomrule
\end{tabular}
\caption{PGIR on the Gemma backbone, mirroring Table~\ref{tab:pgir-application}: all
systems share the same backbone (Gemma\,4 E4B instruction-tuned, greedy decoding) and
the same retriever (E5 over Wikipedia-2018, $k{=}3$), evaluated on the identical
splits as the Qwen table. Cells are exact-match\,/\,F1; Macro is the six-set mean and
Gen is generations per query. The trained systems (SKR, Adaptive-RAG, Probing-RAG,
PGIR) are retrained for this backbone, and PGIR's acceptance threshold and budget are
re-tuned on the dev split only. The confidence-control family ranking is stable
across backbones: PGIR leads macro EM and F1 with Probing-RAG second on both, and
PGIR again trades individual sets to specialists (SeaKR on HotpotQA at 17.5
generations per query, Probing-RAG on TriviaQA). Per column and metric,
\textbf{bold}~$=$~best and \underline{underline}~$=$~second.}
\label{tab:gemma-pgir}
\end{table*}


\end{document}